\pdfoutput=1
\documentclass[10pt,journal]{IEEEtran}
\usepackage{amsmath,amsfonts}

\usepackage{multirow}
\usepackage{booktabs}
\usepackage[ruled,linesnumbered]{algorithm2e}
\usepackage[caption=false,font=normalsize,labelfont=sf,textfont=sf]{subfig}
\usepackage{algorithmic}
\usepackage{enumitem}
\usepackage{hyperref}
\usepackage{graphicx}
\usepackage{cite}
\usepackage{color}

\begin{document}

\title{Erasing Self-Supervised Learning Backdoor by Cluster Activation Masking}

\author{
        Shengsheng~Qian,
        Dizhan~Xue,
        Yifei~Wang,
        Shengjie~Zhang, 
        Huaiwen Zhang, and~Changsheng~Xu,~\IEEEmembership{Fellow,~IEEE}
\IEEEcompsocitemizethanks{

  \IEEEcompsocthanksitem Shengsheng~Qian, Dizhan~Xue, and Changsheng~Xu are with State Key Laboratory of Multimodal Artificial Intelligence Systems, Institute of Automation, Chinese Academy of Sciences, Beijing 100190, China, and also with University of Chinese Academy of Sciences (e-mail: shengsheng.qian@nlpr.ia.ac.cn;
  xuedizhan17@mails.ucas.ac.cn; csxu@nlpr.ia.ac.cn).

  Yifei~Wang is with Xiaomi Inc.,Beijing 100085, China (e-mail: wangyifei15@xiaomi.com).

  Shengjie Zhang is with State Key Laboratory of CNS/ATM, Beihang University, Beijing 100191, China (e-mail: shengjiezhang@buaa.edu.cn).

  Huaiwen~Zhang is with the College of Computer Science, Inner Mongolia University, China (e-mail: huaiwen.zhang@imu.edu.cn).

\textit{This work has been submitted to the IEEE for possible publication. Copyright may be transferred without notice, after which this version may no longer be accessible.}}
\thanks{
(Corresponding author: Changsheng Xu.)}
}

\markboth{Journal of \LaTeX\ Class Files,~Vol.~O, No.~O, February~2024}%
{Shell \MakeLowercase{\textit{et al.}}: A Sample Article Using IEEEtran.cls for IEEE Journals}


\maketitle

\begin{abstract}
    Self-Supervised Learning (SSL) is an effective paradigm for learning representations from unlabeled data, such as text, images, and videos.
    However, researchers have recently found that SSL is vulnerable to backdoor attacks. 
    The attacker can embed hidden SSL backdoors via a few poisoned examples in the training dataset and maliciously manipulate the behavior of downstream models.
    To defend against SSL backdoor attacks, a feasible route is to detect and remove the poisonous samples in the training set.
    However, the existing SSL backdoor defense method fails to detect the poisonous samples precisely.
    In this paper, we propose to erase the SSL backdoor by cluster activation masking and propose a novel PoisonCAM method.
    After obtaining the threat model trained on the poisoned dataset, our method can precisely detect poisonous samples based on the assumption that masking the backdoor trigger can effectively change the activation of a downstream clustering model.
    In experiments, our PoisonCAM achieves 96\% accuracy for backdoor trigger detection compared to 3\% of the state-of-the-art method on poisoned ImageNet-100.
    Moreover, our proposed PoisonCAM significantly improves the performance of the trained SSL model under backdoor attacks compared to the state-of-the-art method.
    Our code, data, and trained models will be open once this paper is accepted.
\end{abstract}


\section{Introduction}
    In recent years, Self-Supervised Learning (SSL) \cite{jaiswal2020survey, chen2024hyperspectral, wang2023self, yang2024ldre, yang2024semantic} has emerged as a powerful paradigm in machine learning, enabling models to learn from vast amounts of unlabeled data. Without the reliance on manual feature engineering or human annotations, SSL is able to learn meaningful representations from unlabeled data and facilitate a range of downstream tasks, such as clustering and classification \cite{he2020momentum, MOCO, Chen2021mocov3, Grill2020byol, 2020SimCLR}. 
    However, recent work \cite{carlini2021poisoning, saha2022backdoor} has found that SSL is vulnerable to backdoor attacks where an attacker can inject a stealthy backdoor trigger into SSL models by poisoning a small number of training samples.

    SSL backdoor attacks \cite{carlini2021poisoning} pose a significant challenge to the security and robustness of SSL models, of which the procedure can be summarized as follows:
    First, an attacker selects a stealthy backdoor trigger for a specific target category.
    Next, the attacker injects the trigger patch into some data of the target category.
    After training on the poisoned dataset, the SSL model will construct a strong correlation between the trigger and the target category.
    Finally, the attacker can manipulate the behavior of the downstream model by attaching the trigger to the input, such as forcing the downstream classifier to misclassify the image as the target category.
    Meanwhile, the attacked model can behave similarly to unattacked models when the trigger is absent in the input, making the injected backdoor stealthy.
    Facing the threat of SSL backdoor attacks, this work aims to erase the SSL backdoor and train a trustworthy SSL model.
    For practicability, we assume that the defender has no prior knowledge of the trigger or target category, and lacks access to trusted data, following \cite{pang2023backdoor, Tejankar2023patchsearch, mu2023progressive, liu2023beating, qi2023towards}.

\begin{figure*}[t]
        \centerline{\includegraphics[width=0.9\linewidth]{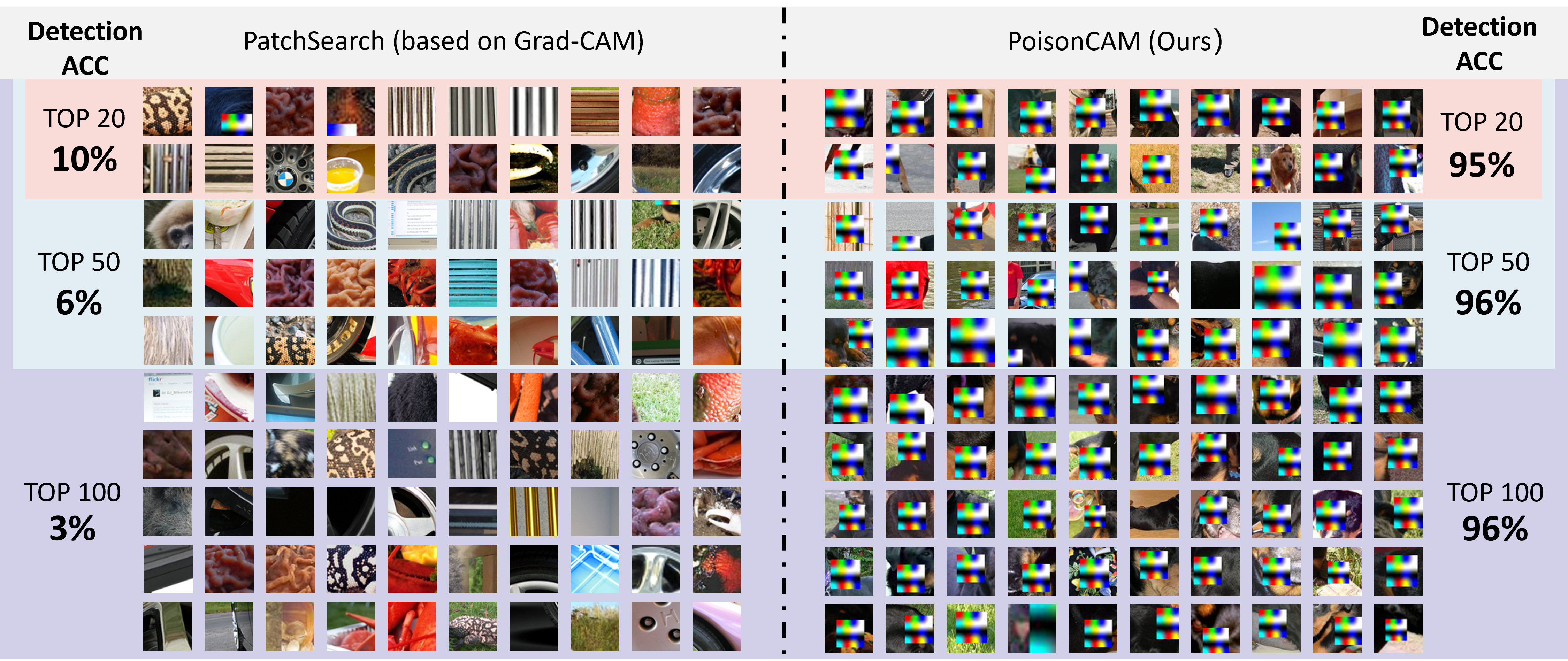}}
        \caption{Retrieved backdoor trigger patches from the poinsoned ImageNet-100 (poison rate 0.5\%, target category ``rottweiler") by PatchSearch (left) \cite{Tejankar2023patchsearch} and our PoisonCAM (right).}
        \label{fig1:cand_trigger}
    \end{figure*}

    To defend against SSL backdoor attacks, a feasible and straightforward route is to detect and remove the poisonous samples in the training set.
    However, since semantic annotations are unavailable in SSL, detecting the backdoor trigger is not a trivial problem and should be achieved in a totally unsupervised manner.
    The existing method \cite{Tejankar2023patchsearch} (i.e., PatchSearch) utilizes Grad-CAM \cite{Selvaraju2017Gradcam} on a downstream clustering model to retrieve trigger patches injected into the dataset, based on which a poison classifier is trained to classify poisonous or clean data.
    However, as shown in Figure \ref{fig1:cand_trigger}, the accuracy of top 20, top 50, and top 100 trigger patches retrieved by PatchSearch are only 10\%, 6\%, and 3\% on the poisoned ImageNet-100.
    Subsequently, PatchSearch is limited in distinguishing between trigger patches and benign patches, reducing the effectiveness of the finally trained SSL models.
    Therefore, this work focuses on addressing a major challenge in SSL backdoor defense: How to accurately retrieve the SSL backdoor trigger patches injected in a poisoned and unlabeled dataset?

    In this paper, we propose PoisonCAM, which aims at accurately detecting poisonous samples in a poisoned and unlabeled dataset to erase the SSL backdoor.
    To retrieve the trigger patches injected into the poisoned dataset, we propose a novel Cluster Activation Masking method.
    Our method is based on the assumption that masking a successful trigger in an image will change the cluster assignment from the target category of the trigger to the true category of the image.
    Therefore, the trigger patches can be detected by comparing the cluster activation under a few random masks.
    As shown in Figure \ref{fig1:cand_trigger}, our PoisonCAM significantly improves the top 20, 50, and top 100 accuracy of the retrieved trigger patches from 10\%, 6\%, and 3\% of the state-of-the-art PatchSearch to 95\%, 96\%, and 96\% on the poisoned ImageNet-100.
    Based on the more accurate trigger patches, our trained poison classifier also significantly improves the precision of detecting poisonous samples from 5.4\% to 49.3\% while achieving even higher recall (see details in Table \ref{table:ssl_remove}).
    Finally, after accurately detecting and removing the poisonous samples, the trained SSL model on the clean-up dataset achieves new state-of-the-art results in terms of both the performance and the backdoor defense.

    In brief, the contributions of this work are summarized as follows:

    \begin{itemize}
        \item 
        We propose PoisonCAM, a novel SSL backdoor defender model, to accurately detect and remove poisonous data from a poisoned and unlabeled dataset. Subsequently, an effective and backdoor-free SSL model can be trained on the clean-up dataset.
        \item 
        We propose a Cluster Activation Masking method to accurately retrieve trigger patches injected into the poisoned dataset. Based on the retrieved trigger patches, an effective poison classifier is trained to classify poisonous or clean data in the training set.
        \item 
        Extensive experimental results on ImageNet-100 and STL-10 demonstrate that our proposed PoisonCAM significantly outperforms the state-of-the-art method for defending against SSL backdoor attacks.
    \end{itemize}

\section{Related Work}
\textbf{Self-Supervised Learning.} 
    The objective of self-supervised learning (SSL) \cite{wu2018unsupervised, yuan2020self, wang2023self2, he2024enhancing, wei2024exploring} is to acquire representations from uncurated and unlabeled data through a pretext task that is derived from the data itself, without any human annotations.
    MoCo \cite{he2020momentum, MOCO, Chen2021mocov3} is a widely employed contrastive SSL algorithm that involves classifying two augmented versions of the same image as positive pairs, which are then contrasted to negative pairs consisting of augmentations from different images. 
    Contrastive SSL algorithms are further explored in \cite{2020SimCLR, Caron2020swav, you2020graph, chuang2020debiased, shah2022max}. 
    BYOL \cite{Grill2020byol} is a non-contrastive SSL algorithm that predicts the target network representation of the same image under a different augmented view without negatives. Non-contrastive SSL algorithms are further developed in \cite{kingma2013auto, yang2019xlnet, Koohpayegani_2021_ICCV, chen2021exploring, kenton2019bert}.
    Despite the remarkable potential, SSL algorithms are not immune to vulnerabilities. 
    In this paper, we study the defense against backdoor attacks on SSL to promote trustworthy and reliable SSL models.

\textbf{SSL Backdoor Attacks.} 
    The purpose of SSL backdoor attacks is to inject a stealthy backdoor trigger into SSL models by poisoning training data, which can be activated to manipulate the behavior of downstream models at test time \cite{jia2022badencoder, tao2023distribution, wang2024badagent}.
    Saha et al. \cite{saha2022backdoor} propose backdoor attacks towards SSL models by attaching a trigger patch to images of a target category. At test time, the downstream classifier has high accuracy on clean images but misclassifies images with the trigger as the target category. The authors also propose a distillation-based defender, which requires a substantial amount of trusted data and may be infeasible in real-world scenarios.
    Li et al. \cite{li2023embarrassingly} recently propose a similar SSL backdoor attack method. The key difference is that they adopt a frequency domain based spectral trigger to resist the data augmentations.
    Carlini and Terzis \cite{carlini2021poisoning} propose backdoor attacks towards CLIP \cite{radford2021learning}, a multimodal contrastive SSL model. Their attack is implemented by injecting triggers into the victim images and tampering with the paired textual captions.

    \textbf{SSL Backdoor Defenses.} Despite the prosperity of backdoor defense for supervised learning \cite{huang2021backdoor, zhu2023neural, zhang2023backdoor, min2023towards, xu2023medic}, SSL backdoor defense is a relatively under-explored and more challenging problem.
    Tejankar et al. \cite{Tejankar2023patchsearch} explore to defend against patch-based SSL backdoor attacks \cite{saha2022backdoor}. Their method adopts Grad-CAM \cite{Selvaraju2017Gradcam} on a clustering model to detect candidate triggers and train a poison classifier to identify and delete poisonous samples in the training data.
    However, they compromisingly delete a large amount of clean samples due to the low accuracy of their poison classifier.
    Bansal et al. \cite{bansal2023cleanclip} propose to defend the multimodal SSL backdoor attacks on CLIP. 
    They find that simply integrating an intra-modal contrastive loss can effectively mitigate multimodal SSL backdoor attacks.
    \textbf{In this paper, we follow the research line of patch-based SSL backdoor attacks on visual SSL models \cite{saha2022backdoor, Tejankar2023patchsearch} due to their substantial feasibility and destructiveness.}
    We propose a novel PoisonCAM method based on Cluster Activation Masking to detect and delete the poisonous samples in the training set while deleting as few clean samples as possible.

	\begin{figure*}[t]
		\centerline{\includegraphics[width=0.95\linewidth]{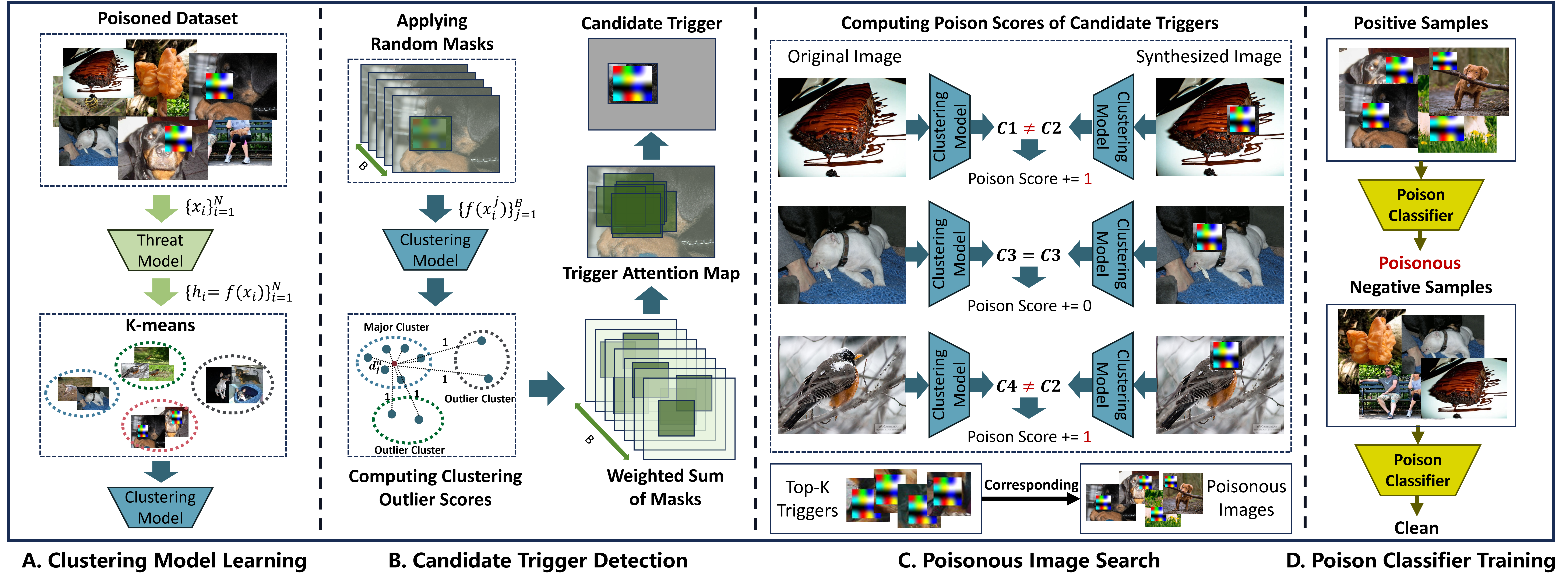}}
		\caption{The overview of PoisonCAM: (1) Learn a clustering model on the poisoned dataset by $k$-means algorithm; (2) Detect the candidate trigger in each image based on clustering outlier scores of random masks and the weighted sum of masks as the trigger attention map; 
		(3) Compute the poison scores of candidate triggers and retrieve the top-k triggers with corresponding poisonous images; (4) Train a poison classifier to identify and remove poisonous samples in the poisoned dataset.}
		\label{framework}
	\end{figure*}
 
\section{Threat Model}
    We introduce the threat model under the SSL backdoor attack proposed in \cite{saha2022backdoor}.
    The primary objective of the attacker is to manipulate the output of a downstream model which is constructed based on the SSL model.
    In this paper, we take a downstream image classifier as the example, following previous work \cite{saha2022backdoor, Tejankar2023patchsearch}.
    Assume that an unlabeled poisoned dataset $\boldsymbol{X} = \{x_i \in \mathbb{R}^{C \times H \times W}\}_{i=1}^{N}$ contains $N$ images where ${x}_i$ is the $i$-th image, $C$ denotes the number of channels, $H$ and $W$ denote the height and the width of an image, respectively. 
    The attacker's purpose is two-fold. Firstly, the attacker aims to surreptitiously implant a backdoor into an SSL model, enabling the downstream classifier to misclassify an incoming image as the target category if the image contains the attacker-designated trigger $t$. Secondly, the attacker seeks to hide the backdoor's presence by ensuring that the performance of the downstream classifier is similar to that of a classifier based on an unattacked SSL model when the trigger is absent.
    The attacker's objectives can be achieved through a technique known as ``data poisoning". This involves the attachment of a small trigger patch to a few images of the preselected target category. The attacker posits that the SSL algorithm associates the trigger with the target category, resulting in a successful attack. 
    Then, the model $f(\cdot)$ trained on the poisoned dataset $\boldsymbol{X}$ via SSL algorithms (e.g., MoCo \cite{he2020momentum, MOCO, Chen2021mocov3}) is defined as the threat model.
    Moreover, the learned feature $\boldsymbol{h}_i = f(x_i)$ can be utilized in the downstream tasks.

\section{Proposed Defender Model}
The objective of the defender model is to remove the SSL backdoor, eliminating the hidden correlation between the attacker-designed trigger and the target category. 
Simultaneously, the defender should avoid impairing the model's performance on clean data.
To enhance the practicability, we assume that the defender should achieve this without any prior knowledge of the trigger or target category, and lacking access to trusted data, following \cite{pang2023backdoor, Tejankar2023patchsearch, mu2023progressive, liu2023beating, qi2023towards}.

In this section, we propose our defender model named PoisonCAM, which aims at identifying the poisonous samples in the training set and filtering them out to form a clean-up training dataset $\bar{\boldsymbol{X}}\subset \boldsymbol{X}$. 
%
%
The overall architecture of our method is shown in Figure \ref{framework}, which mainly consists of four steps: 
(1) Learning clustering model for the poisoned dataset ${\boldsymbol{X}}$;
(2) Detect the candidate triggers for all images in ${\boldsymbol{X}}$;
(3) Search for the top-$k$ poisonous samples with the highest poison scores;
(4) Train a poison classifier to find and delete all poisonous samples in ${\boldsymbol{X}}$.
Subsequently, the clean-up training dataset $\bar{\boldsymbol{X}}$ can be formed and a backdoor-free SSL model can be trained on $\bar{\boldsymbol{X}}$.

\subsection{Clustering Model Learning} \label{clustering}
    Due to the lack of labels in SSL, we first learn a clustering model $C(\cdot)$ for features $\{f(x_i)\}_{i=1}^N$ by $k$-means algorithm \cite{MacQueen1967SomeMF} to capture the semantics of training data, as follows: 
    \begin{equation}
        y_i = C(f(x_i)),
        \label{eq:kmeans}
    \end{equation}
    where $y_i \in \{1,...,l\}$ represents the corresponding cluster label and $l$ is a hyper-parameter.
    Since the threat model is sensitive to triggers, the poisonous images with triggers will tend to be classified into a cluster of triggers.
    The clustering model $C(\cdot)$ is fixed and will be utilized in the following.

\subsection{Candidate Trigger Detection for Images}\label{perturbing}

\begin{figure}[t]  
        \begin{minipage}{2.5cm}
            \centerline{\includegraphics[width=2.2 cm,height=2.2 cm]{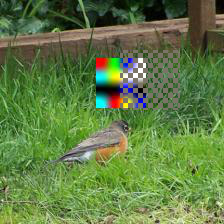}}
            \centerline{\footnotesize{(a) 0-1 Interval}}
        \end{minipage}
        \hfill
        \begin{minipage}{2.5cm}
            \centerline{\includegraphics[width=2.2 cm,height=2.2 cm]{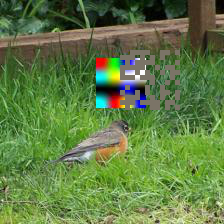}}
            \centerline{\footnotesize{(b) Random}}
        \end{minipage}
        \hfill
        \begin{minipage}{2.5cm}
            \centerline{\includegraphics[width=2.2 cm,height=2.2 cm]{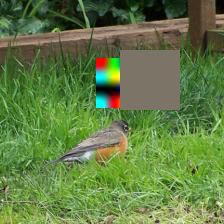}}
            \centerline{\footnotesize{(c) Full Coverage}}
        \end{minipage}
        \vfill
        \caption{An example of three masking strategies.}
        \label{fig:sample_mask}
    \end{figure} 
    In candidate trigger detection, we aim to detect the candidate trigger region in every image $x_i$.
    The state-of-the-art PatchSearch \cite{Tejankar2023patchsearch} employs Grad-CAM \cite{Selvaraju2017Gradcam} to detect the pivotal region of an image $x_i$ for clustering as the candidate trigger region. 
    However, previous work \cite{Chattopadhay2018campp, Jiang2021layercam, Wang2020scorecam} has found that Grad-CAM may fail to locate the pivotal region for the downstream tasks. Moreover, as we empirically show the detection results on poisoned ImageNet-100 in Figure \ref{fig1:cand_trigger} (a), Grad-CAM suffers from detecting the trigger in this poisoned SSL dataset.
    Therefore, we propose a novel Cluster Activation Masking method based on the assumption that masking a trigger $t$ in an image $x_i$ will change the cluster assignment $y_i$ from the cluster of triggers to the true cluster of $x_i$.
    By analyzing the cluster activation under $B$ different random masks, we can locate the candidate trigger $t_i$ in $x_i$.
    
    Specifically, using the obtained clustering model $C(\cdot)$ in Equation \ref{eq:kmeans}, we aim at locating the region of the potential trigger in $x_i$.
    For any image $x_i\in{\boldsymbol{X}}$, we randomly initialize $B$ masks $\boldsymbol{M} = \{\boldsymbol{m}_j\in \{0,1\}^{H \times W}\}_{j=1}^{B}$.
    In particular, we set the pixel value to zero if the corresponding mask value is 1 and keep the pixel value unchanged if the corresponding mask value is 0.
    After randomly selecting a masking window of the size of $[w, w]$, we have empirically tested three masking strategies, as shown in Figure \ref{fig:sample_mask}.
    Although it seems that 0-1 Interval Masking and Random Masking can break the pattern of the trigger while protecting the semantics of benign regions, we empirically found in experiments (see Section \ref{sec:ablation}) that Full Coverage Masking is the optimal strategy.
    We apply masks $\{\boldsymbol{m}_j\}_{j=1}^B$ to obtain $B$ degraded images $\boldsymbol{D}=\{x_i^j=x_i \odot (1-\boldsymbol{m}_j)\}_{j=1}^{B}$, where $\odot$ denotes the element-wise multiplication. 
    Then we classify the features of $\boldsymbol{D}$ by the clustering model $C(\cdot)$ to get cluster labels $\{\eta_j=C(f(x_i^j))\}_{j=1}^{B}$  and the distances to the assigned cluster centers $\{d_j\}_{j=1}^{B}$. 
    Since the size of the backdoor trigger is much smaller than the whole image, most masks cannot change the clustering assignment of $x_i^j$. Therefore, most samples from $\boldsymbol{D}$ are assigned to the same cluster and we denote this cluster as $\eta^*$. 
    To scale cluster distances, we utilize Max-Min Scaling to normalize $d_j$ as $d^n_j = \frac{d_j-\min_k d_k}{\max_k d_k-\min_k d_k}\in[0,1]$. Then, we compute the clustering outlier score $a_j$ of the mask $\boldsymbol{m}_j$ as follows:
    \begin{equation}
        \begin{aligned}
        a_j = \begin{cases}
            d_j^n & {\eta}_j=\eta^*\\
            1 &{\eta}_j\ne \eta^* .
            \end{cases}
        \end{aligned} 
        \label{eq:weight}
    \end{equation}
    
    For a poisonous sample $x_i$, $\eta^*$ should be the target category of trigger $t$.
    ${\eta}_j\ne\eta^*$ means the trigger is successfully masked. When the trigger is not successfully masked,  larger $d_j^n$ to the cluster center $\eta^*$ means more region of the trigger is masked.
    Therefore, $a_j$ can represent the importance score of successfully masking the trigger.
    Based on the scores $\{a_i\}_{i=1}^B$, the masks are aggregated through weighted summation to obtain an attention map of the trigger, as follows:
    \begin{align}
        A = \frac{\sum_{j= 1}^{B} \boldsymbol{m}_j \times {a}_j}{\sum_{j= 1}^{B} \boldsymbol{m}_j} \in \mathbb{R}^{H \times W},
        \label{eq:attention}
    \end{align}
    where $A$ is the attention map of the trigger, and the value $A_{mn}$ positively correlates to the presence possibility of the trigger.
    After obtaining the attention map of the image $x_i$, we select the window $t_i$ of size $w \times w$ with the largest sum attention value in the window.
    Therefore, $t_i$ can be regarded as the optimal window of the potential trigger $t$ in $x_i$ and we denote $t_i$ as the candidate trigger for simplicity.

\subsection{Poisonous Image Search}\label{PClearning}
    In this section, we aim to find a set of highly poisonous images in $\boldsymbol{X}$.
    To quantify the poisonousness of a given image $x_i$, we define the poison score of $x_i$.
    Our method is based on the assumption that pasting a real trigger onto an image will strongly change its clustering assignment while the effectiveness of pasting a benign region is much weaker.
    Specifically, we first obtain a fixed test set $\boldsymbol{X}^f$ by sampling a few images per cluster that are closest to their respective cluster centers.
    Subsequently, we paste the candidate trigger $t_i$ onto all images in $\boldsymbol{X}^f$ and get their new cluster assignments.
    Finally, the poison score of $x_i$ is calculated as the number of images in $\boldsymbol{X}^f$ whose cluster assignments are changed after pasting the candidate trigger $t_i$.
    
    To find a few highly poisonous images, we can compute the poison scores of all images and simply take the top-$k$ image as the poisonous samples. 
    However, to improve the efficiency and utilize the clustering information of images, we adopt a heuristic search strategy following \cite{Tejankar2023patchsearch} to iteratively compute poison scores for a part of (not all) images in $\boldsymbol{X}$.
    At every iteration, we first compute the poison scores of $s$ randomly selected samples per cluster and use the sums to represent the poison scores of clusters. 
    Since clusters with lower poison scores are less likely to be poisoned, we then remove a fraction $r$ of the clusters with the least poison scores and will not compute poison scores for images in these clusters.
    After several iterations, all clusters are removed, and a part of images in $\boldsymbol{X}$ are scored during the iterative procedure.
    Finally, we take the top-$k$ scored images with the highest poison scores to form a poison set $\boldsymbol{X}^p$. 
    This poison set will be utilized to train a classifier of poisonous samples.
    In this heuristic algorithm, the number of scored images is unfixed.
    This poison set will be utilized to train a classifier of poisonous samples.

\begin{table*}[ht]
\centering
\caption{
\textbf{Defense results.} We compared the performance of the attacked SSL models with different defense methods under various attack settings on the validation set. ACC denotes accuracy. ASR denotes attack success rate. Clean Val denotes the clean validation set and Poisoned Val denotes the poisoned validation set.}  

\setlength\tabcolsep{4pt}{
\begin{tabular}{c|c|cccc|cccc|cccc}
\hline
\multicolumn{2}{c|}{{Attack Settings}}  & \multicolumn{4}{c|}{w/o Defense}  & \multicolumn{4}{c|}{PatchSearch}  & \multicolumn{4}{c}{PoisonCAM}  \\ \hline 
\multirow{2}{*}{Dataset}    & \multirow{2}{*}{Target Category}
& \multicolumn{2}{c}{Clean Val} & \multicolumn{2}{c|}{Poisoned Val} 
& \multicolumn{2}{c}{Clean Val} & \multicolumn{2}{c|}{Poisoned Val} 
& \multicolumn{2}{c}{Clean Val} & \multicolumn{2}{c}{Poisoned Val} \\ \cline{3-14} 
&  & ACC$\uparrow$    & ASR$\downarrow$    & ACC$\uparrow$   & ASR$\downarrow$    & ACC$\uparrow$   & ASR$\downarrow$   & ACC$\uparrow$   & ASR$\downarrow$    & ACC$\uparrow$   & ASR$\downarrow$      & ACC$\uparrow$   & ASR$\downarrow$ \\ \hline
\multirow{11}{*}{\begin{tabular}[c]{@{}c@{}}ImageNet-100\\ (poison rate 0.5\%)\end{tabular}}
& rottweiler        & 69.4     & 0.5      & 27.7    & 63.9  & 68.0  & 0.5   & 61.9 & 0.4    & 70.6   & 0.5  & 65.2     & 0.5    \\
& tabby cat         & 69.1     & 0.0      & 30.2    & 61.8  & 67.3  & 0.0   & 60.7 & 0.1    & 70.9   & 0.1  & 64.7     & 0.1    \\
& ambulance         & 69.4     & 0.0      & 57.3    & 9.6   & 66.6  & 0.0   & 56.5 & 3.5    & 69.4   & 0.1  & 62.3     & 0.1    \\
& pickup truck      & 70.1     & 0.3      & 58.5    & 9.3   & 65.6  & 0.4   & 60.2 & 0.3    & 71.0   & 0.3  & 65.4     & 0.3    \\
& laptop            & 69.2     & 0.9      & 38.0    & 52.4  & 66.2  & 1.0   & 48.9 & 24.5   & 68.9   & 0.8  & 61.6     & 2.4    \\
& goose             & 69.4     & 0.2      & 44.7    & 35.5  & 66.8  & 0.2   & 61.1 & 0.3    & 70.0   & 0.2  & 63.4     & 0.3    \\
& pirate ship       & 69.5     & 0.0      & 52.5    & 22.2  & 66.0  & 0.1   & 51.8 & 15.1   & 68.5   & 0.1  & 61.5     & 0.2    \\
& gas mask          & 68.6     & 0.3      & 33.4    & 58.8  & 69.4  & 1.1   & 63.1 & 2.1    & 68.8   & 0.9  & 63.3     & 2.1    \\
& vacuum cleaner    & 69.1     & 1.1      & 44.0    & 32.2  & 67.8  & 1.2   & 61.0 & 1.1    & 70.0   & 1.4  & 62.7     & 1.2    \\
& American lobster & 68.8     & 0.1      & 44.0    & 42.5  & 65.2  & 0.3   & 59.6 & 0.3    & 69.2   & 0.2  & 62.4     & 0.2    \\ \cline{2-14}
& average           & 69.3     & \textbf{0.3}      & 43.0    & 38.8  & 66.9  & 0.5   & 58.5 & 4.8    & \textbf{69.7}   & 0.4  & \textbf{63.2}     & \textbf{0.7}    \\ \hline
\multirow{11}{*}{\begin{tabular}[c]{@{}c@{}}ImageNet-100\\ (poison rate 1.0\%)\end{tabular}}                                               
& rottweiler        & 69.4     & 0.4      & 26.2    & 70.9    & 67.1   & 0.5  & 60.9   & 0.5   & 69.3   & 0.7   & 63.5  & 0.6  \\
& tabby cat         & 69.3     & 0.0      & 25.8    & 69.9    & 68.0   & 0.5  & 62.8   & 0.7   & 68.7   & 0.5   & 63.4  & 0.6  \\
& ambulance         & 69.5     & 0.0      & 49.4    & 23.3    & 67.0   & 0.2  & 58.1   & 2.3   & 69.9   & 0.2   & 62.4  & 0.4  \\
& pickup truck      & 69.2     & 0.3      & 52.6    & 22.4    & 67.6   & 0.4  & 61.8   & 0.4   & 69.0   & 0.4   & 63.2  & 0.5  \\
& laptop            & 69.0     & 0.8      & 31.6    & 61.4    & 69.0   & 1.1  & 61.5   & 3.4   & 69.4   & 0.9   & 62.7  & 1.3  \\
& goose             & 69.7     & 0.2      & 40.0    & 47.8    & 61.9   & 0.4  & 56.5   & 0.4   & 68.3   & 0.5   &  61.7  & 0.6   \\
& pirate ship       & 69.0     & 0.1      & 49.1    & 30.8    & 68.8   & 0.5  & 60.9   & 4.0   & 68.9   & 0.4   & 61.8  & 0.8  \\
& gas mask          & 68.7     & 0.3      & 29.2    & 65.4    & 68.2   & 1.3  & 60.4   & 4.3   & 69.1   & 1.3   & 62.5  & 2.3  \\
& vacuum cleaner    & 68.9     & 1.0      & 39.2    & 44.5    & 69.0   & 1.3  & 61.0   & 2.1   & 69.9   & 1.4   & 63.3  & 1.4  \\
& American lobster  & 69.0     & 0.1      & 27.2    & 68.1    & 67.9   & 0.7  & 61.7   & 1.9   & 69.9   & 0.6   & 62.5  & 2.1 \\  \cline{2-14}
& average           & \textbf{69.2}     & \textbf{0.3}      & 37.0    & 50.4    & 67.4   & 0.7  & 60.6   & 2.0   & \textbf{69.2}   & 0.7   &  \textbf{62.7}  & \textbf{1.1}    \\ \hline
\multirow{4}{*}{\begin{tabular}[c]{@{}c@{}}STL-10\\ (poison rate 5.0\%)\end{tabular}}
& bird              & 63.2     & 5.1      & 50.8    & 28.7    & 62.6   & 3.1  & 56.2   & 7.1   & 65.4   & 4.9   & 61.0  & 5.1  \\
& car               & 62.0     & 1.8      & 41.2    & 41.3    & 61.8   & 1.8  & 40.8   & 3.3   & 62.8   & 1.1   & 43.6  & 3.1  \\
& deer              & 65.8     & 2.9      & 43.8    & 25.1    & 64.0   & 3.3  & 42.2   & 4.8   & 63.8   & 4.0   & 51.0  & 4.2 \\  \cline{2-14}
& average           & 63.7     & 3.3      & 45.3    & 31.7    & 62.8   & \textbf{2.7}  & 46.4   & 5.1   & \textbf{64.0}   & 3.3   &  \textbf{51.9}  & \textbf{4.1}    \\ \hline
\end{tabular}}

\label{table:ssl_compare} 
\end{table*}

\subsection{Poison Classifier Training}
    Aiming at precisely detecting all poisonous samples in $\boldsymbol{X}$, we train a ResNet \cite{ResNet} as the poison classifier.
    Specifically, for every $x_i\in\boldsymbol{X}$, we randomly select a sample $x_k$ in the poison set $\boldsymbol{X}^p$ and paste its candidate trigger $t_k$ at a random location on $x_i$.
    These synthesized images as the positive samples and original images in $\boldsymbol{X}$ as the negative samples form the poison classification set $\Tilde{\boldsymbol{X}}$.
    Since poisonous samples from $\boldsymbol{X}$ are noisy for poison classification, we eliminate a proportion $p$ of images with the highest poison scores in $\boldsymbol{X}$ and utilize strong augmentations and early stopping to alleviate the noisy problem.
    Then, the poison classifier trained on $\Tilde{\boldsymbol{X}}$ is applied to $\boldsymbol{X}$, and all images classified as ``poisonous" are removed to form a clean-up training dataset $\bar{\boldsymbol{X}}$. 
    Finally, after erasing the SSL backdoor in the poisoned dataset, we can train a benign SSL model on $\bar{\boldsymbol{X}}$, which can achieve similar performance to the threat model $f(\cdot)$ on clean samples but will not be manipulated by the attacker-designated trigger.

\section{Experiments}
    Focusing on the self-supervised learning (SSL) backdoor defense task, we conduct extensive experiments on two widely-adopted benchmark datasets.

\subsection{Datasets}

    Following previous work \cite{Tejankar2023patchsearch}, we adopt \textbf{ImageNet-100} \cite{Tian2020imagenet100}, which contains images belonging to 100 randomly sampled classes from the 1000 classes of ImageNet \cite{Olga2015imagenet}. The training set has about 127K samples and the validation set has 5K samples. 
    We also adopt \textbf{STL-10} \cite{AdamCoates2011AnAO} that contains 500/800 training/validation images for each of the 10 classes.

\subsection{Attack Setting} 
    Following the SSL backdoor attacks proposed in \cite{saha2022backdoor}, we randomly adopt 10 different target categories and trigger patches. 
    In every experiment, we set a single target category and use a single trigger patch.
    On ImageNet-100, we set the poisoning rates as $0.5\%$ and $1.0\%$, which mean $50 \%$ and $100 \%$ images of the target category being poisoned to form a poisonous dataset. 
    On STL-10, we set the poisoning rates as $5.0\%$, which mean $50 \%$ images of the target category being poisoned. 
    The HTBA triggers \cite{saha2020hidden, sun2020poisoned} are pasted with size $50 \times 50$ onto the images and $25\%$ of the edges is reserved on four sides of the images. 
    After injecting the triggers, we train a threat model on the poisoned dataset $\boldsymbol{X}$. 
    Following PatchSearch \cite{Tejankar2023patchsearch}, we adopt ViT-B \cite{dosovitskiy2020vit} as the backbone with MoCo-v3 algorithm \cite{Chen2021mocov3} to train threat models on poisoned datasets.

\subsection{Baseline Method and Evaluation Metrics} 
    We adopt the state-of-the-art SSL defense method PatchSearch \cite{Tejankar2023patchsearch} and the naive method without defense as the baseline methods.
    %
    We fairly compare our proposed PoisonCAM with baselines by sharing the weights of the threat model.
    For the evaluation, we train a linear classifier on a trained SSL model by randomly sampling a $1.0\%$ subset from the clean labeled training dataset, following PatchSearch. 
    Since PatchSearch has not released its trained models, we reimplement PatchSearch using the official code. \footnote{https://github.com/UCDvision/PatchSearch}.
    We will release our poisoned datasets, code, and trained models once this paper is accepted.

    We denote the original validation set of ImageNet-100 as the clean validation set, which consists of 50 randomly selected images from each category, and form a poisoned validation set by randomly pasting the attacked trigger on the images of the clean validation set.
    Similarly, we construct the clean and poisoned validation sets of STL-10.
    We then assess the results on both the clean validation set and the poisoned validation set by using the following metrics: Accuracy (ACC) and Attack Success Rate (ASR) \cite{yao2019latent, zhu2020gangsweep, li2020invisible} for the attacked class. 
    Attack Success Rate (ASR) refers to the proportion of non-target classes misclassified as the targeted class.
    All results are averaged over 5 independent runs with different seeds.

\begin{table}[ht]
\centering
\caption{
\textbf{Poison detection results.} We compare the average results over target categories for detecting poisonous images in the poisoned datasets. \textit{Total Rem.} denotes the number of total removed samples.}  
\begin{tabular}{c|c|c|c}
\hline
Dataset                                                                                     & Metric     & PatchSearch & PoisonCAM \\ \hline
\multirow{3}{*}{\begin{tabular}[c]{@{}c@{}}ImageNet-100\\ (poison rate 0.5\%)\end{tabular}} & Total Rem. & 12807.5     & 1570.8    \\
                                                                                            & Recall$\uparrow$     & 84.4        & \textbf{98.7}      \\
                                                                                            & Precision$\uparrow$  & 5.4         & \textbf{49.3}      \\ \hline
\multirow{3}{*}{\begin{tabular}[c]{@{}c@{}}ImageNet-100\\ (poison rate 1.0\%)\end{tabular}} & Total Rem. & 10740.7     & 2214.9    \\
                                                                                            & Recall$\uparrow$     & 99.0        & \textbf{99.1}      \\
                                                                                            & Precision$\uparrow$  & 20.9        & \textbf{64.3}      \\ \hline
\multirow{3}{*}{\begin{tabular}[c]{@{}c@{}}STL-10\\ (poison rate 5.0\%)\end{tabular}}       & Total Rem. & 744.0       & 332.0     \\
                                                                                            & Recall$\uparrow$     & 98.5        & \textbf{99.1}      \\
                                                                                            & Precision$\uparrow$  & 62.5        & \textbf{78.6}      \\ \hline
\end{tabular}
\label{table:ssl_remove} 
\end{table}

\section{Implementation Details}
    We utilize PyTorch \cite{paszke2019pytorch} to implement all experiments on four GeForce RTX 3090 GPUs. 
    We employed the following parameters for PoisonCAM on ImageNet-100 as same as PatchSearch: the cluster count of $l=1000$, the number of the flip test set is set as $1000$, samples per cluster $s=2$, and removal of $r=25\%$ of candidate clusters after each iteration. The search window size was set to $w=60$, using a complete sampling approach. This method typically involves searching through approximately 8,000 images on ImageNet-100.
    For training the poison classifier, we sample top-20 poisonous images (i.e., $|\boldsymbol{X}^p|=20$) and remove the top $10\%$ poisonous samples in $\boldsymbol{X}$ to reduce noise in the poison classification set $\bar{\boldsymbol{X}}$. 
    Besides, we set the number of masks as $B=256$.

    Given the height $H$ and width $W$ of the images, the $(x,y)$ positions of our sampled masks are in cyclic groups. For a pixel $(x,y)$ in the sampled mask, its real position is $(x\ {\rm mod}\ H, y\ {\rm mod}\ W)$, where the originally sampled positions $x$ and $y$ can exceed $H$ and $W$.

\subsection{Results and Discussions}

The results on poisoned ImageNet-100 and STL-10 under different attack settings are shown in Table \ref{table:ssl_compare}. Based on the results, we have the following observations:
    \begin{itemize}
        \item 
        Our proposed PoisonCAM significantly outperforms baselines against SSL backdoor attacks. 
        Specifically, on poisoned validation sets, PoisonCAM achieves average ACC improvements of $4.7\%$, $2.1\%$, and $5.5\%$ and average ASR reductions of $4.1\%$, $0.9\%$, and $1.0\%$ on ImageNet-100 (poison rate 0.5\%), ImageNet-100 (poison rate 1.0\%), and STL-10 (poison rate 5.0\%), respectively. 
        These results indicate that PoisonCAM can better defend against backdoor attacks by accurately removing poisonous samples for detoxified training.
        \item 
        Our PoisonCAM achieves similar or even higher ACC than w/o Defense on clean validation sets where backdoor triggers are absent.
        Compared to PatchSearch, PoisonCAM achieves $2.8\%$, $1.8\%$, and $1.2\%$ average ACC improvements on ImageNet-100 (poison rate 0.5\%), ImageNet-100 (poison rate 1.0\%), and STL-10 (poison rate 5.0\%), respectively while achieving similar ASRs.
        This is because PoisonCAM can classify poisonous samples more accurately and delete fewer benign samples to facilitate sufficient training.
        Since PatchSearch deletes too many clean samples, its ACC obviously declines compared to w/o Defense on three datasets.
    \end{itemize}

\begin{figure}[ht]  
    \begin{minipage}{0.49\linewidth}
        \centerline{\includegraphics[width=\linewidth]{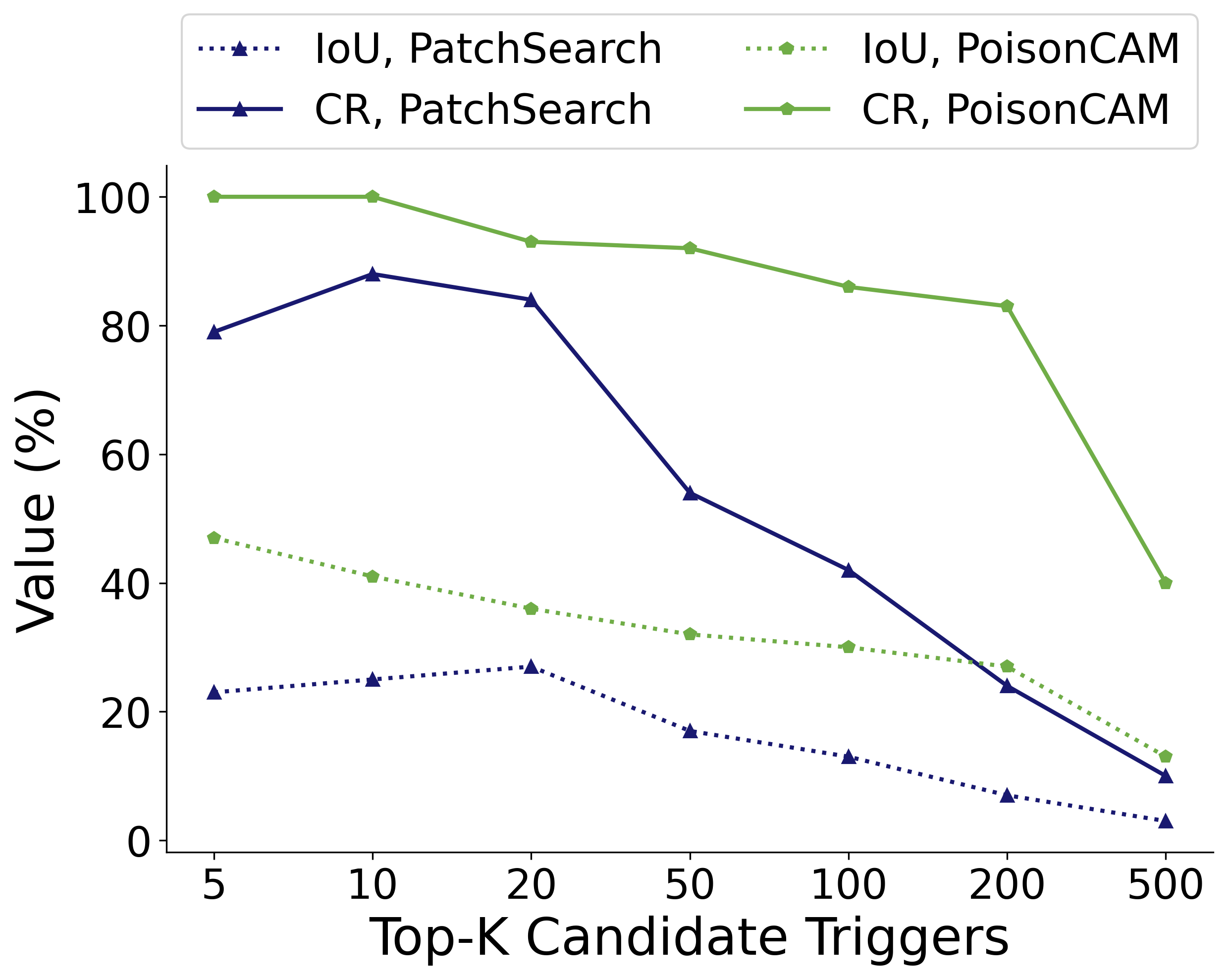}}
        \centerline{\footnotesize{(a) 2-target attack}}
    \end{minipage}
    \hfill
    \begin{minipage}{0.49\linewidth}
        \centerline{\includegraphics[width=\linewidth]{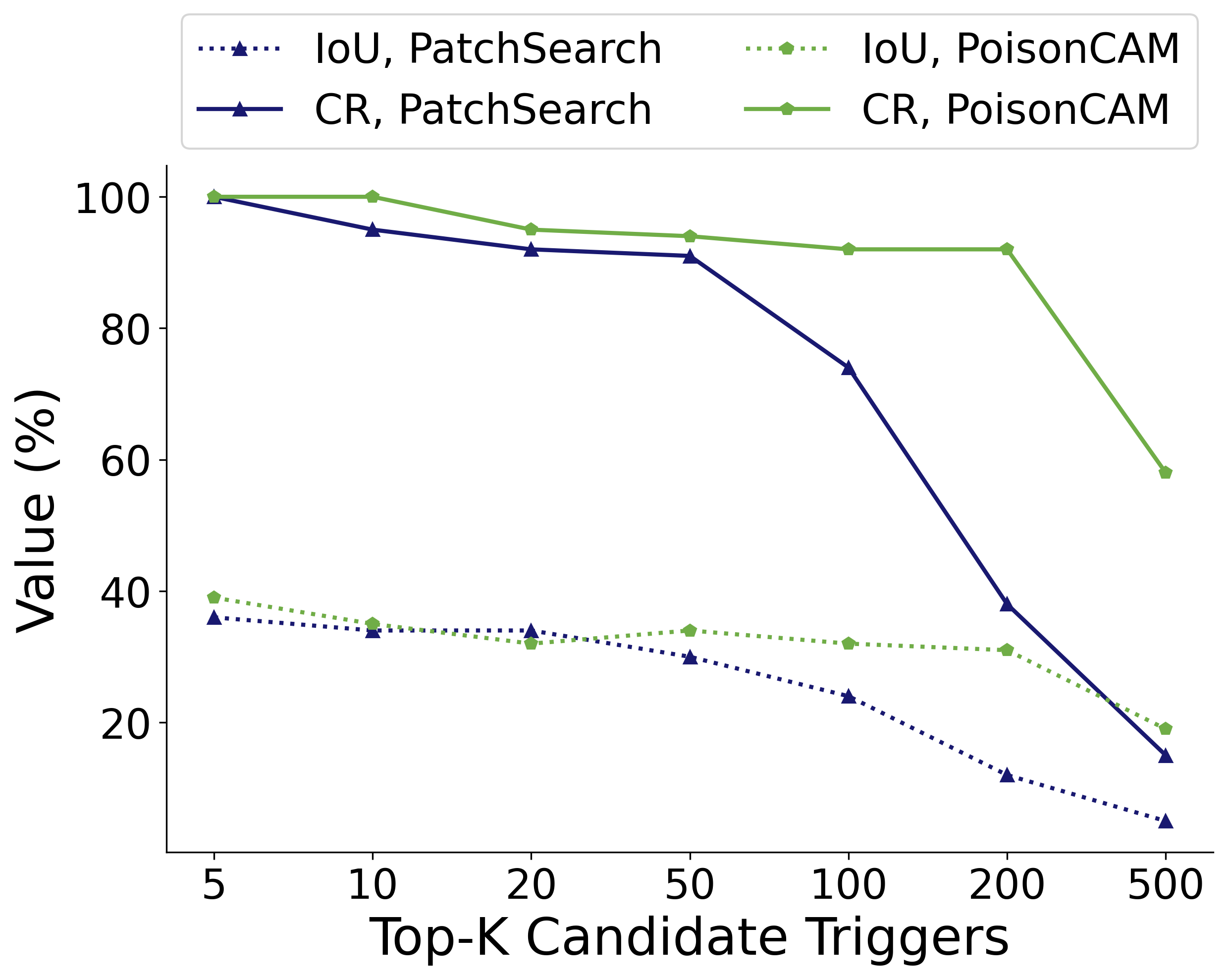}}
        \centerline{\footnotesize{(b) 3-target attack}}
    \end{minipage}  
    \vfill
    \caption{Results of the detected top-$k$ candidate triggers by different methods against multi-target attacks.}
    \label{fig:mixed_attack}
\end{figure} 

\begin{figure}[ht]  
    \begin{minipage}{0.49\linewidth}
        \centerline{\includegraphics[width=\linewidth, height=3.3cm]{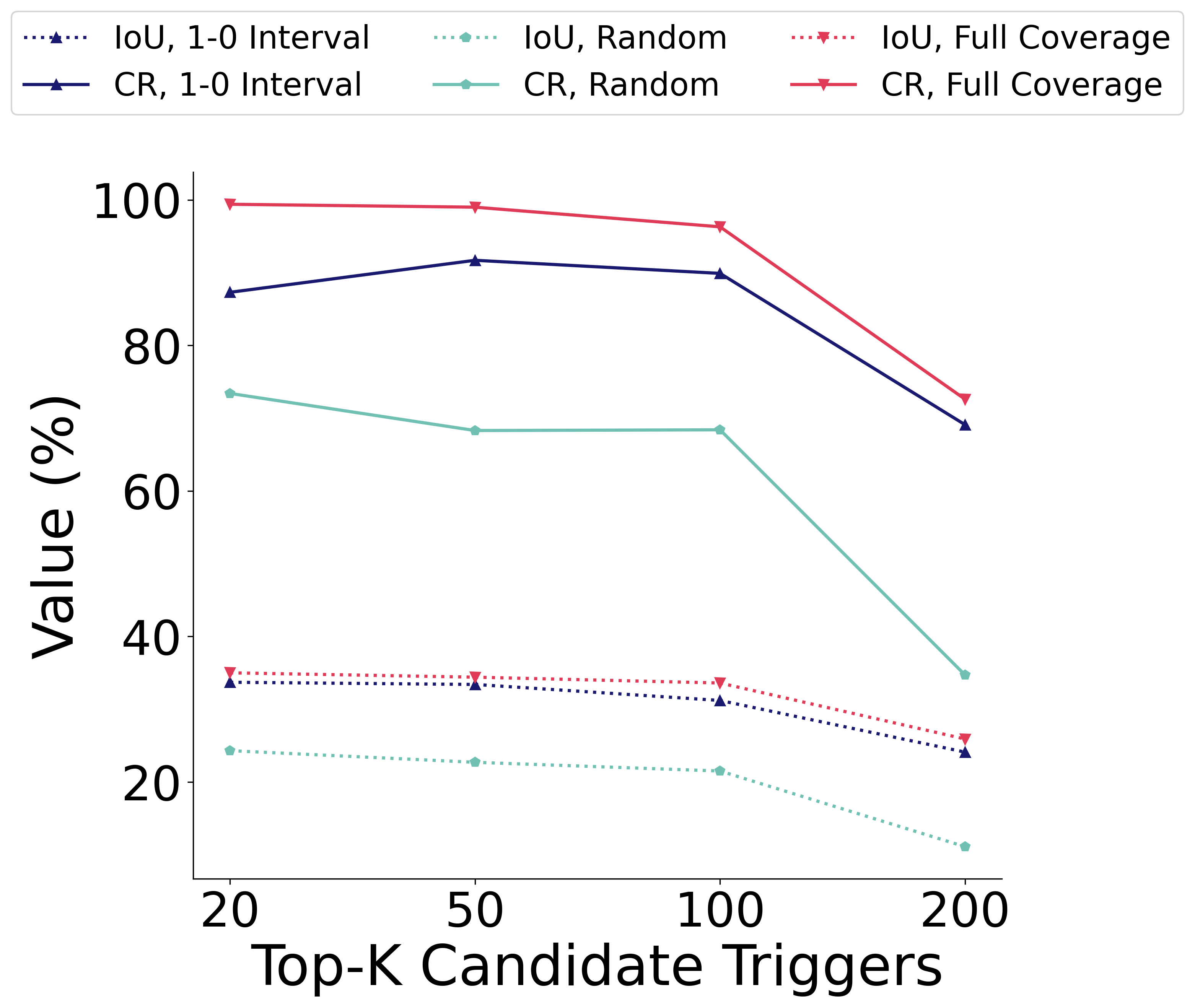}}
        \centerline{\footnotesize{(a) IoU \& CR}}
    \end{minipage}
    \hfill
    \begin{minipage}{0.49\linewidth}
        \centerline{\includegraphics[width=\linewidth, height=3.3cm]{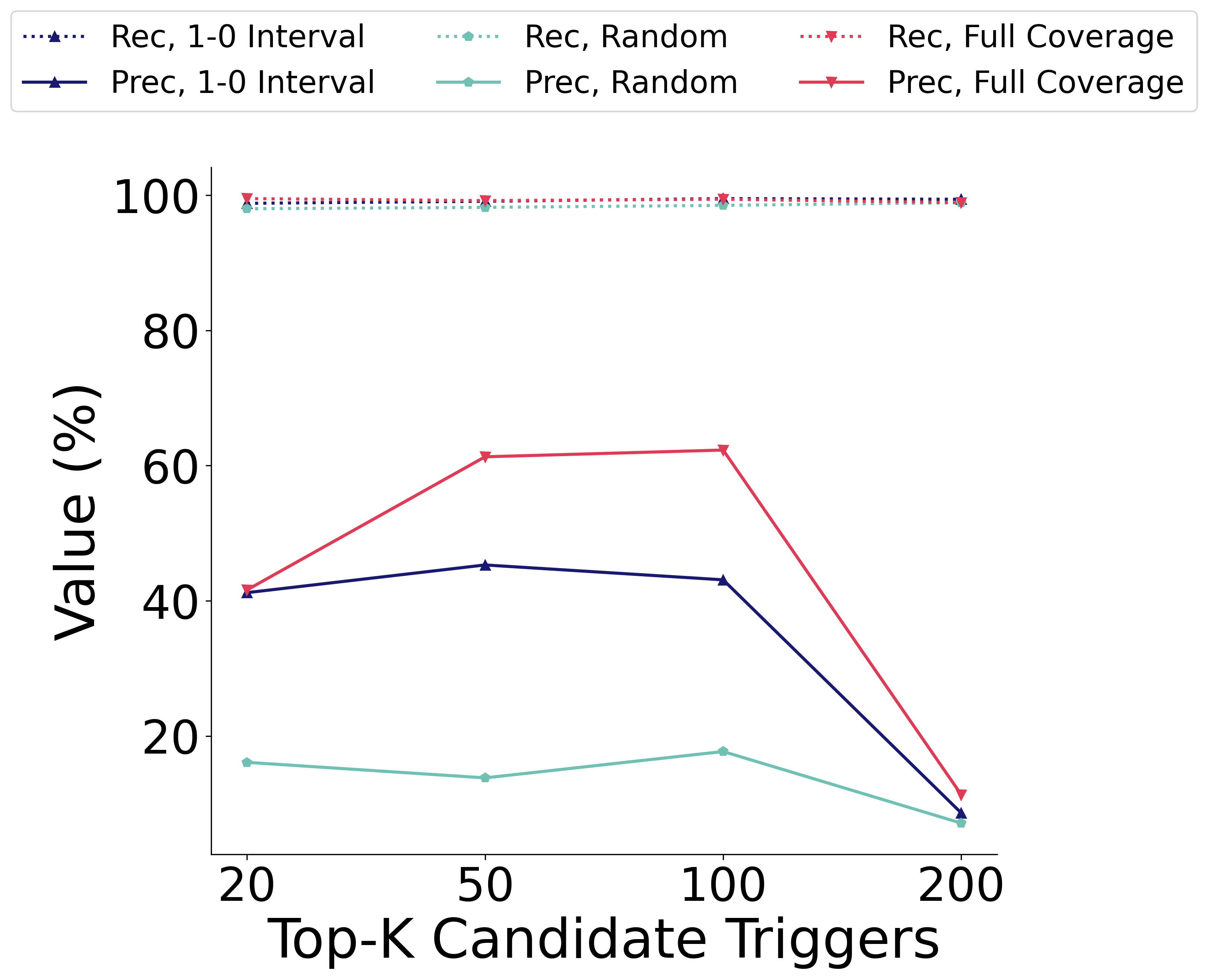}}
        \centerline{\footnotesize{(b) Recall \& Precision}}
    \end{minipage}  
    \vfill
    \caption{Ablation study on different masking strategies on ImageNet-100 (poison rate 0.5\%, target category ``rottweiler"). Rec., Prec. denotes Recall and Precision.}
    \label{fig:abl}
\end{figure} 

    \begin{figure*}[ht]
        \centerline{\includegraphics[width=0.8\linewidth]{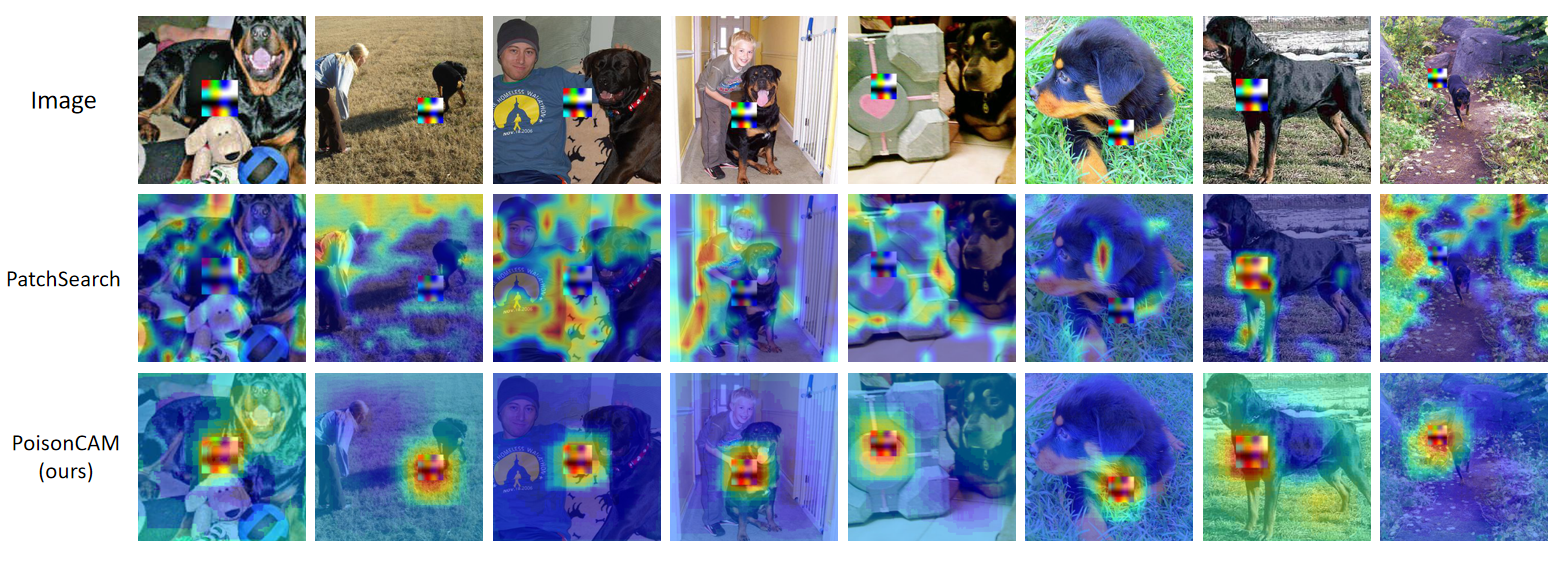}}
        \caption{Visualization of trigger attention maps generated by PatchSearch and our PoisonCAM on ImageNet-100 (poison rate 0.5\%, target category ``rottweiler").}
        \label{fig:Attention}
    \end{figure*} 

    \begin{figure*}[ht]
    \centerline{\includegraphics[width=0.8\linewidth]{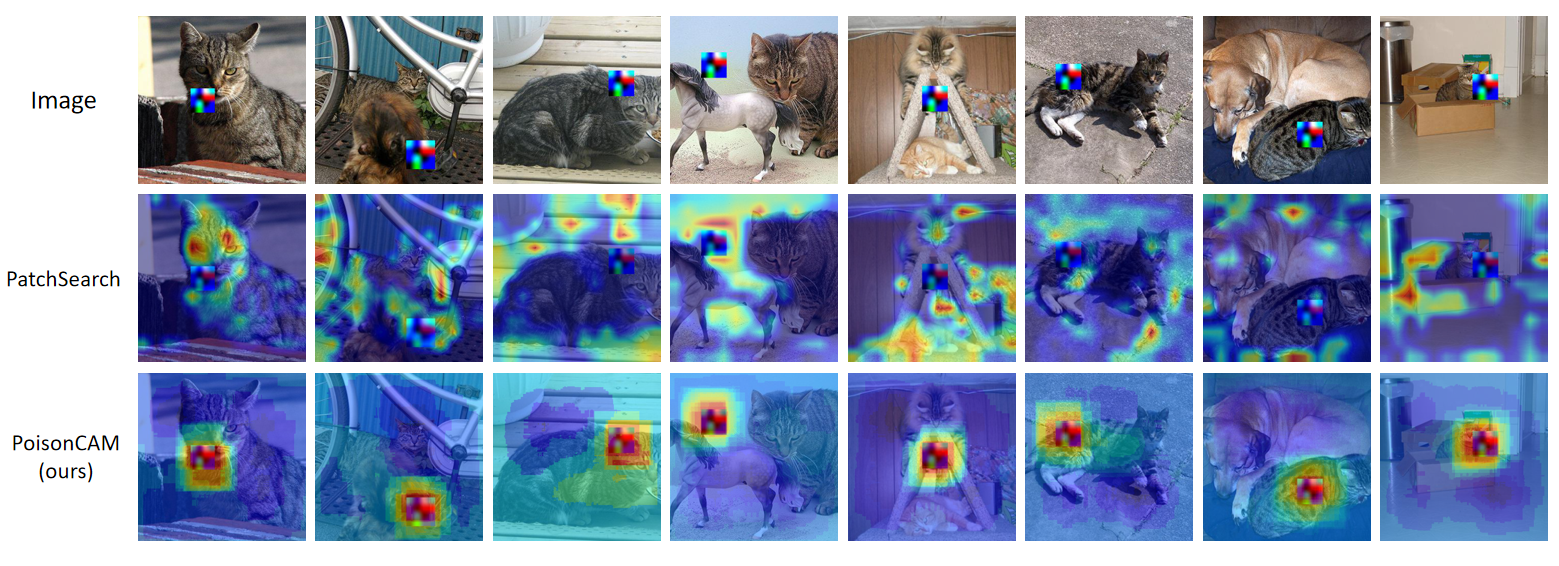}}
    \caption{Visualization of trigger attention maps generated by PatchSearch and our PoisonCAM on ImageNet-100 (poison rate 0.5\%, target category “tabby cat”).}
    \label{fig:Attention_11}
\end{figure*}
\begin{figure*}[ht]
    \centerline{\includegraphics[width=0.8\linewidth]{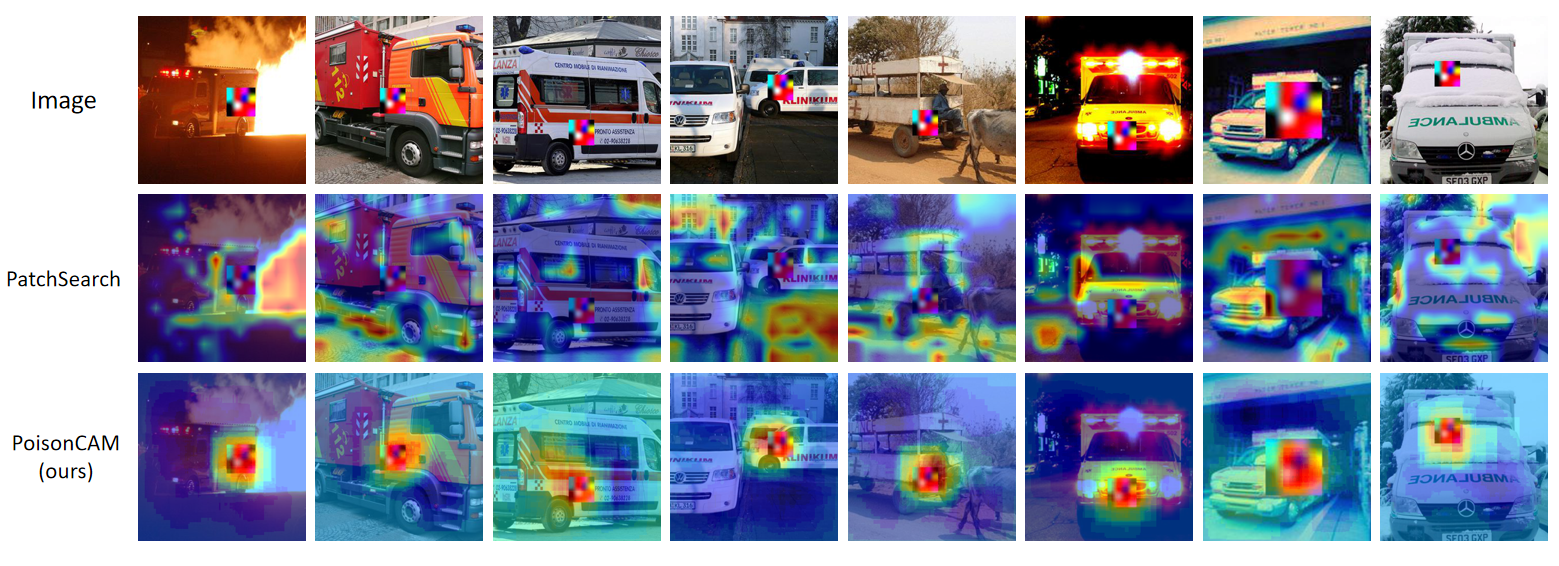}}
    \caption{Visualization of trigger attention maps generated by PatchSearch and our PoisonCAM on ImageNet-100 (poison rate 0.5\%, target category “ambulance”).}
    \label{fig:Attention_12}
\end{figure*} 
\begin{figure*}[ht]
    \centerline{\includegraphics[width=0.8\linewidth]{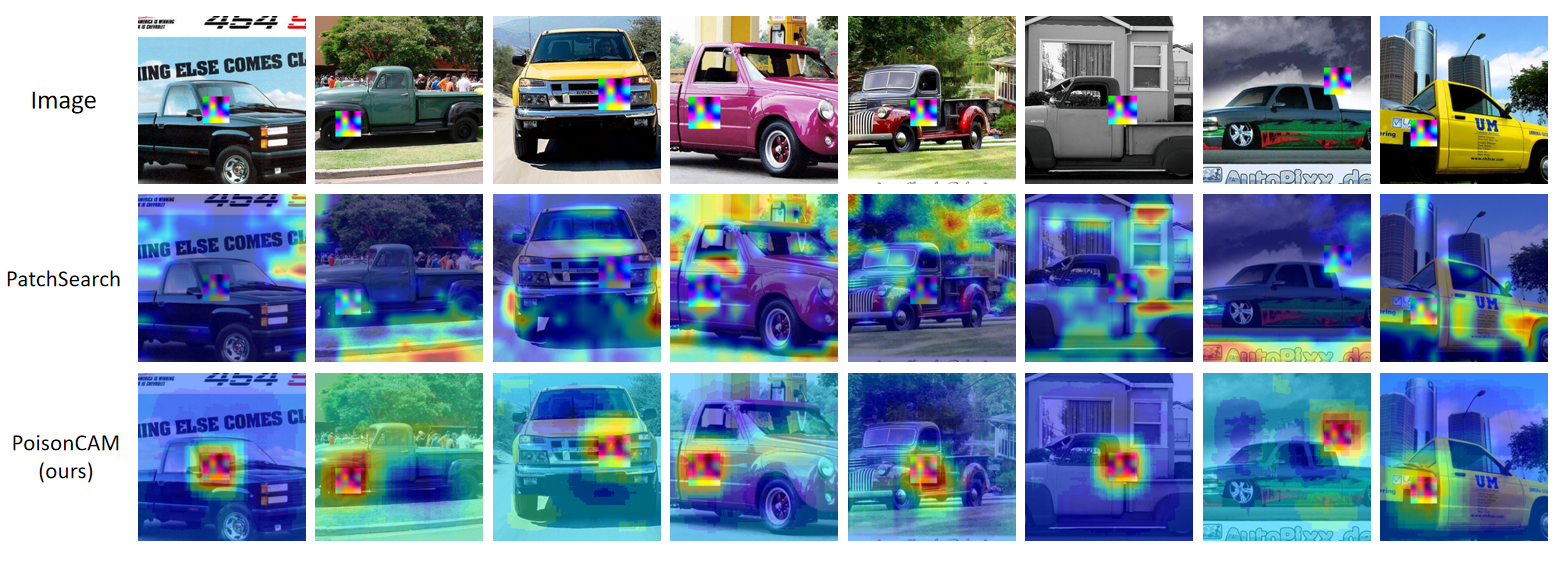}}
    \caption{Visualization of trigger attention maps generated by PatchSearch and our PoisonCAM on ImageNet-100 (poison rate 0.5\%, target category “pickup truck”).}
    \label{fig:Attention_13}
\end{figure*} 

    \begin{figure*}[ht]
    \begin{minipage}{0.49\linewidth}
        \centerline{\includegraphics[width=\linewidth]{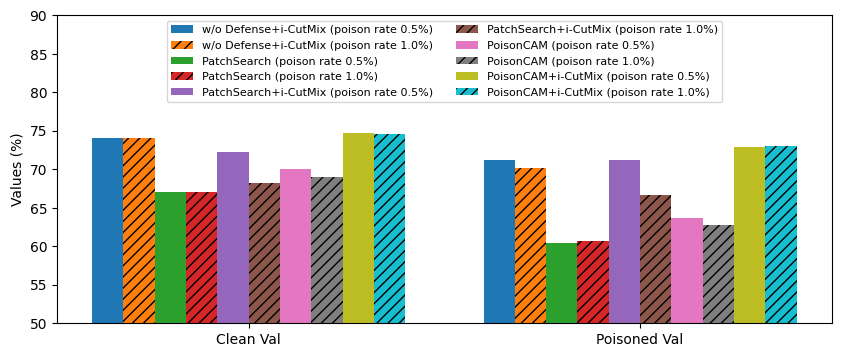}}
        \centerline{\footnotesize{(a) ACC $\uparrow$}}
    \end{minipage}
    \hfill
    \begin{minipage}{0.49\linewidth}
        \centerline{\includegraphics[width=\linewidth]{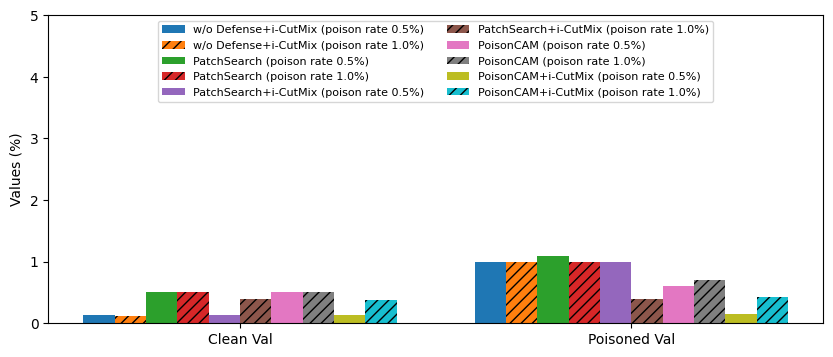}}
        \centerline{\footnotesize{(b) ASR $\downarrow$}}
    \end{minipage}  
    \vfill
    \caption{Defense results with i-CutMix augmentation on ImageNet-100 (poison rates 0.5\% and 1.0\%, target category ``rottweiler"). }
    \label{fig:i-CutMix}
\end{figure*} 
    
\subsection{Analysis of Poisonous Image Detection}
A key step in PatchSearch and our PoisonCAM is to detect poisonous images in the training set and remove them.
To further investigate the effectiveness of our method, we analyze the results of poisonous image detection by PoisonCAM and PatchSearch. In Table \ref{table:ssl_remove}, we report the number of total removed images (Total Rem.), recall, and precision on three datasets. We have the following observations:
\begin{itemize}
    \item
    Compared to PatchSearch, our PoisonCAM significantly improves average precision by 43.9\%, 43.4\%, and 16.1\% on ImageNet-100 (poison rate 0.5\%), ImageNet-100 (poison rate 1.0\%), and STL-10 (poison rate 5.0\%), respectively.
        These results demonstrate that PoisonCAM can detect poisonous images more accurately and reduce mistakenly removed benign samples, which also results in the lower Total Rem.
        Subsequently, PoisonCAM can facilitate more sufficient training to improve the performance of the SSL models. 
    \item
    Moreover, PoisonCAM achieves similar or higher recall compared to PatchSearch on three datasets. 
        Especially, the recall of PoisonCAM is always higher than 95\% on all datasets. 
        As a result, PoisonCAM can remove poisonous samples more effectively and facilitate detoxified SSL training. 
\end{itemize}

We include the detailed results of every target category in Supplementary Materials.

\subsection{Results of Multi-Target Attacks}
To further investigate the effectiveness of different methods, we conduct experiments against multi-target attacks.
Specifically, we utilize multiple target categories and correlate different backdoor triggers to different target categories. 
We combine target categories of ``rottweiler" and ``tabby cat" to conduct 2-target attacks, and further add ``ambulance" to conduct 3-target attacks on ImageNet-100 (poison rate 0.5\%). 
As shown in Figure \ref{fig:mixed_attack}, we report the results of top-$k$ retrieved candidate triggers by PatchSearch and PoisonCAM.
We introduce two metrics, Intersection over Union (IoU) \cite{ren2015faster, xue2024few}, and Catch Rate (CR). IoU measures the union divided by the intersection between the real triggers and the retrieved triggers. CR measures the ratio of the real triggers contained in the retrieved triggers.

We have the following observations:
(1) Our PoisonCAM significantly outperforms PatchSearch against multi-target attacks. For example, PoisonCAM always achieves the best CR with every search number from top-5 to top-500.
These results show that PoisonCAM can locate larger trigger areas than PatchSearch, which facilitates training a more accurate poison classifier to identify poisonous samples.
(2) CR and IoU of both methods decline with the search number increasing from top-5 to top-500. 
However, diverse candidate triggers are also necessary for capturing the global characteristics of the real triggers. 
Therefore, an appropriate search number should be selected to balance the accuracy and diversity of the detected triggers.

\subsection{Ablation Study} \label{sec:ablation}
To further investigate the effectiveness of the designed masking strategies in Section \ref{perturbing}, we design different masking strategies and conduct an ablation study on ImageNet-100 (poison rate 0.5\%) with the target category ``rottweiler":
Recalling that we set the size of masking windows as $[w, w]$, we compare three masking strategies:
    \begin{enumerate}
        \item 
        0-1 Interval Masking: A block is defined with a size of $[w', w']$ $ (w' \ll w)$. 
        The entire window is evenly divided into blocks. 
        Every block is assigned a value of 0 or 1, $s.t.$, every two adjacent blocks are assigned different values, as shown in Figure \ref{fig:sample_mask} (a).
        \item 
        Random Masking \cite{he2022masked}: Again, the entire window is evenly divided into blocks. 
        Half of the blocks are randomly selected and assigned a value of 0, others are assigned a value of 1, as shown in Figure \ref{fig:sample_mask} (b).
        \item 
        Full Coverage Masking: The entire window is assigned a value of 1, as shown in Figure \ref{fig:sample_mask} (c).
    \end{enumerate}
As we report the results in Figure \ref{fig:abl}, we have the following observations:
(1) Full Coverage Masking achieves both the highest IoU and CR among all strategies and can catch nearly the entire ($\sim$100\% CR) trigger regions with the search number from top-20 to top-100. 
(2) Simultaneously, the trained poison classifier with Full Coverage Masking achieves the highest precision and nearly 100\% recall with different search numbers.
Based on these observations, we empirically select Full Coverage Masking in our method.

\begin{figure*}[t]  
    \begin{minipage}{0.23\linewidth}
        \centerline{\includegraphics[width=\linewidth]{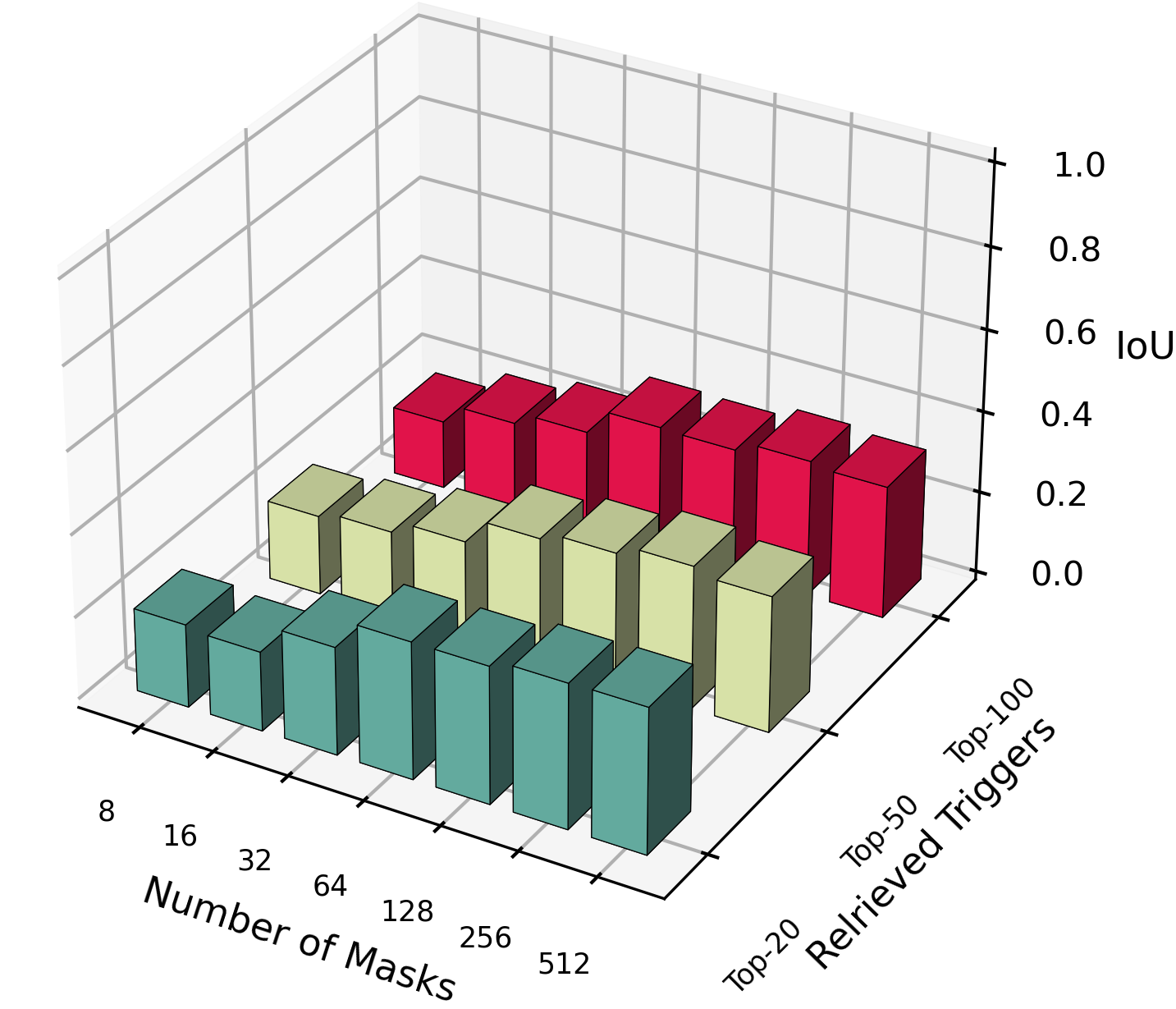}}
        \centerline{\footnotesize{(a) IoU}}
    \end{minipage}
    \hfill
    \begin{minipage}{0.23\linewidth}
        \centerline{\includegraphics[width=\linewidth]{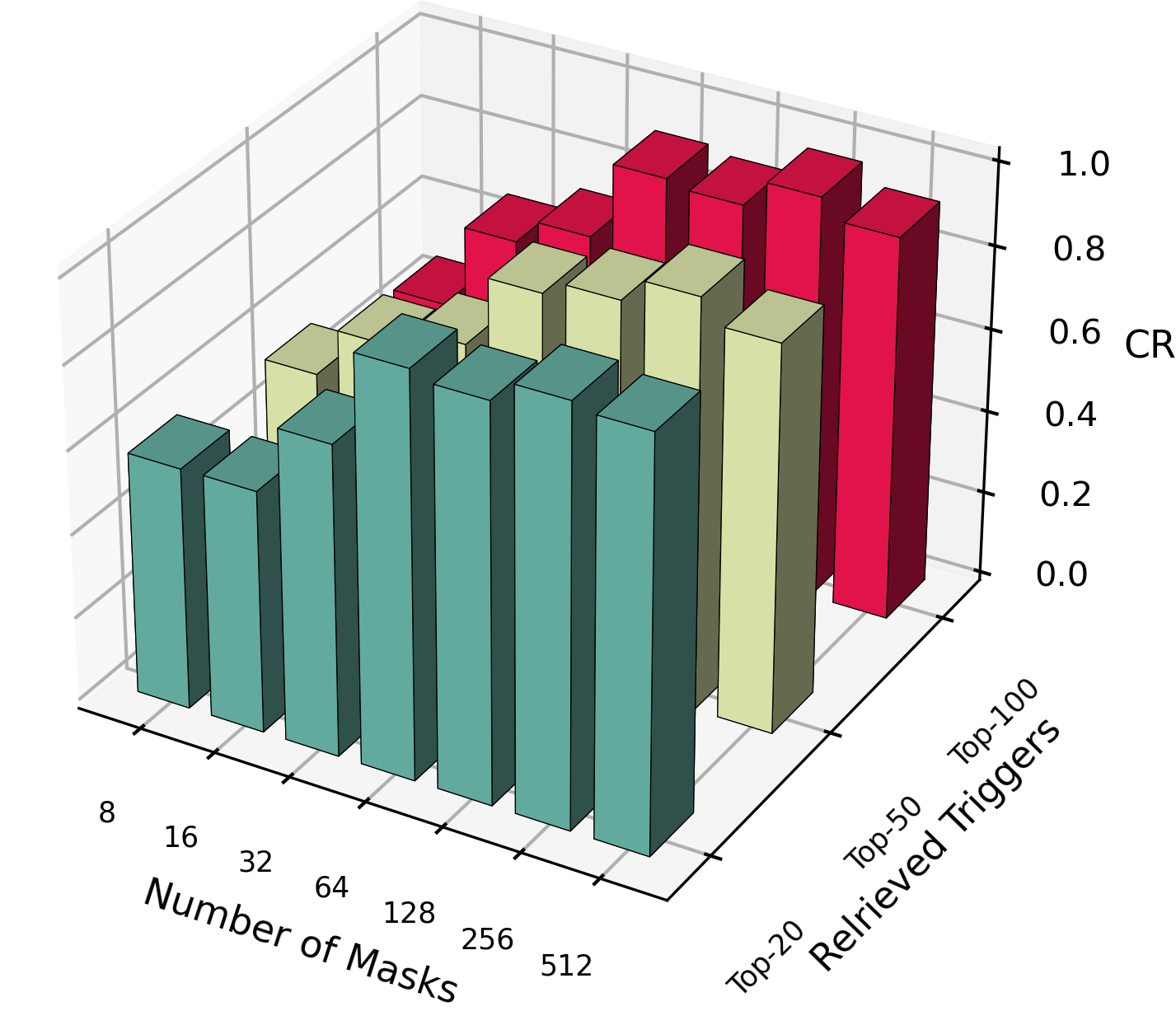}}
        \centerline{\footnotesize{(b) CR}}
    \end{minipage}
    \hfill
    \begin{minipage}{0.23\linewidth}
        \centerline{\includegraphics[width=\linewidth]{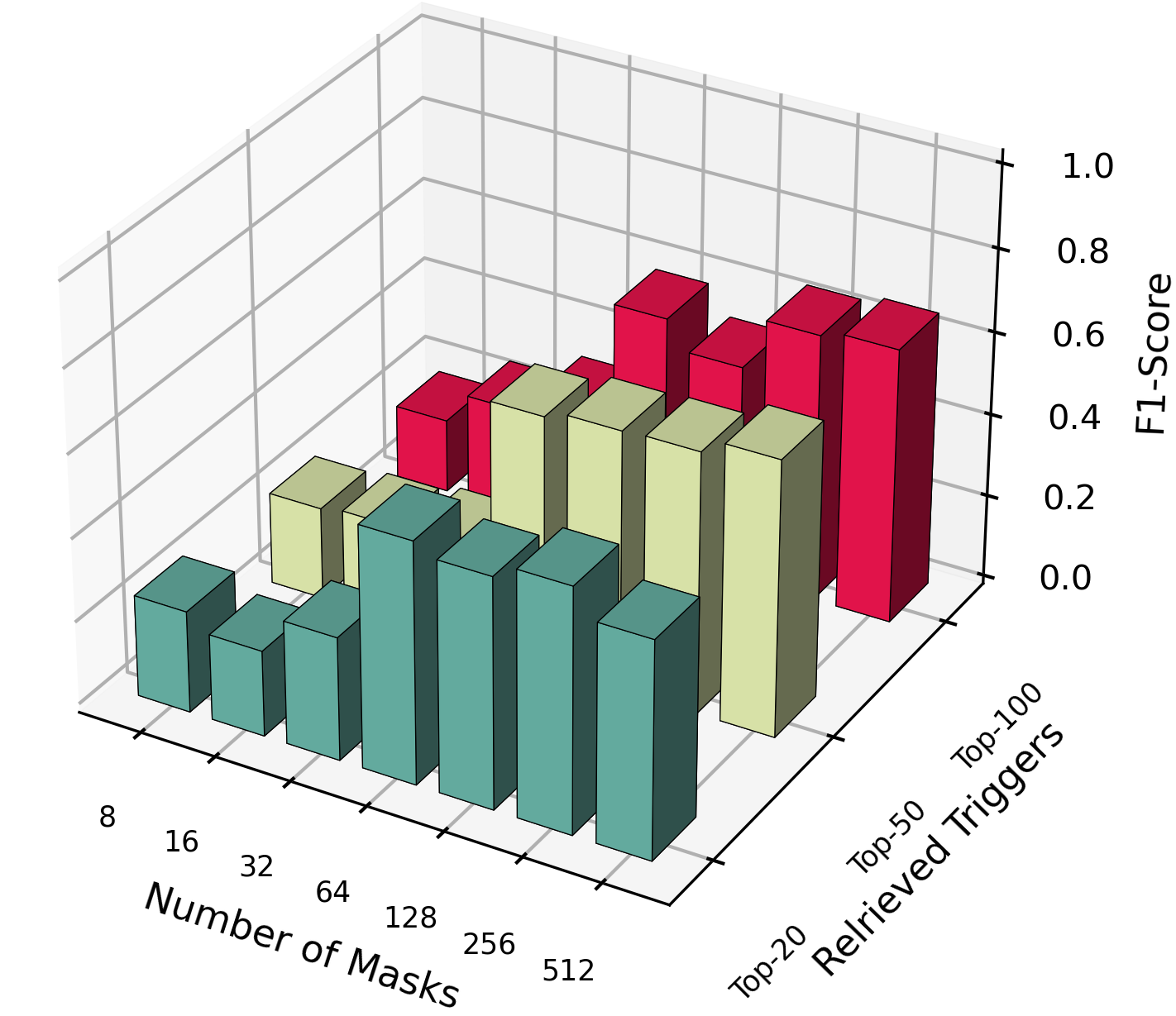}}
        \centerline{\footnotesize{(c) F1-score}}
    \end{minipage}
    \hfill
    \begin{minipage}{0.26\linewidth}
        \centerline{\includegraphics[width=\linewidth]{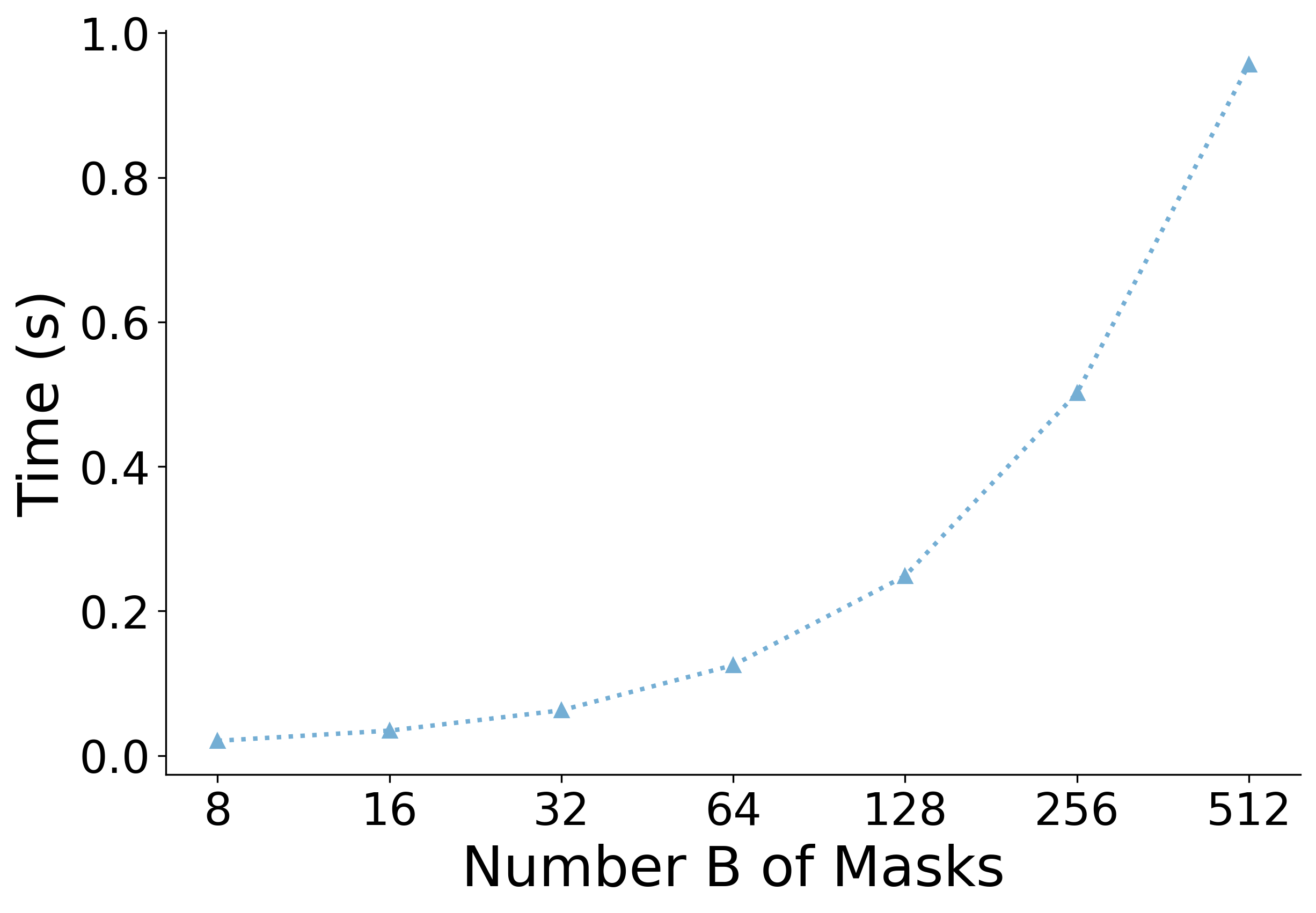}}
        \centerline{\footnotesize{(d) Average Processing Time}}
    \end{minipage}  
    \vfill
    \caption{Hyper-parameter sensitivity study on the number $B$ of masks on ImageNet-100 (poison rate 0.5\%, target category ``rottweiler"). (a-b) IoU and CR of the detected candidate triggers. (c) F1-score of the trained poison classifier. (d) Average processing time per image of candidate trigger detection.}
    \label{fig:para}
\end{figure*} 

\begin{table}[ht]
\centering
\caption{The attack effectiveness throughout the backdoored SSL process.}  
\setlength\tabcolsep{4pt}
\begin{tabular}{c|cccc|cccc}
\hline
\multirow{3}{*}{Epoch} 
& \multicolumn{4}{c|}{laptop}  & \multicolumn{4}{c}{goose} \\
& \multicolumn{2}{c}{Clean Data} & \multicolumn{2}{c|}{Poison Data} & \multicolumn{2}{c}{Clean Data} & \multicolumn{2}{c}{Poison Data} \\ \cline{2-9} 
    & Acc             & FP           & Acc             & FP             & Acc             & FP           & Acc            & FP             \\ \hline
20  & 24.4            & 47           & 18.7            & 114            & 24.5            & 56           & 18.2           & 115            \\
40  & 38.5            & 44           & 31.8            & 416            & 39.0            & 39           & 31.7           & 181            \\
60  & 51.8            & 45           & 27.8            & 2390           & 52.2            & 30           & 41.3           & 566            \\
80  & 59.4            & 44           & 33.3            & 2368           & 59.8            & 17           & 46.7           & 705            \\
100 & 64.1            & 40           & 38.9            & 2238           & 64.2            & 16           & 46.6           & 1223           \\
120 & 67.1            & 39           & 42.4            & 2112           & 66.7            & 12           & 46.0           & 1467           \\
140 & 68.2            & 38           & 40.8            & 2325           & 68.3            & 13           & 47.0           & 1485           \\
160 & 69.0            & 46           & 39.5            & 2492           & 69.0            & 12           & 48.1           & 1420           \\
180 & 69.4            & 43           & 38.6            & 2533           & 69.2            & 10           & 42.9           & 1888           \\
200 & 69.2            & 43           & 38.0            & 2593           & 69.4            & 11           & 44.7           & 1758           \\ \hline
\end{tabular}
\label{tab:poison_epoch}
\end{table}

\subsection{Analysis of Trigger Attention Maps} \label{sec:Attention}
To further investigate the process of two defenders, PoisonCAM and PatchSearch, we conduct a qualitative analysis of the generated trigger attention maps.
Both methods generate trigger attention maps for images and select the hottest regions as the candidate triggers.
PatchSearch adopts Grad-CAM \cite{Selvaraju2017Gradcam} to compute the trigger attention, while we propose a Cluster Activation Masking method in Section \ref{perturbing} to solve this problem.
Therefore, we visualize the cases of trigger attention maps generated by PoisonCAM and PatchSearch on ImageNet-100 (poison rate 0.5\%) with the target category ``rottweiler" in Figure \ref{fig:Attention}.
Our method can always accurately locate the backdoor triggers contained in these images, while PatchSearch focuses on more dispersed and maybe irrelevant regions.
In Figures \ref{fig:Attention_11}, \ref{fig:Attention_12}, \ref{fig:Attention_13}, we visualize more cases of trigger attention maps generated by PoisonCAM and PatchSearch on ImageNet-100 (poison rate $0.5\%$) with the target categories “tabby cat”, “ambulance” and “pickup truck”.
Our method can always accurately locate the backdoor triggers contained in these images, while PatchSearch focuses on more dispersed and maybe irrelevant regions.
These results further demonstrate that our PoisionCAM can significantly improve the detection accuracy of the injected backdoor trigger, which is fundamental for defending against SSL backdoor attacks.

\subsection{Defend with Data Augmentation}
Except for cleaning poisoned data, data augmentations can also relieve backdoor attacks since they may break triggers in training,
Moreover, these two types of methods can be integrated in defending.
Therefore, we conducted defense experiments with complex i-CutMix \cite{lee2020mix} augmentation on datasets containing 0.5\% and 1.0\% poisoned samples. We experiment with i-CutMix, Patch Search, our PoisonCAM, and their combinations.

The experimental results are shown in Figure \ref{fig:i-CutMix}, from which we have the following observations:

\begin{enumerate}
    \item 
    i-CutMix outperforms PatchSearch in ACC and can achieve sufficiently low ASR. This shows that appropriate data augmentation can relieve backdoor attacks on SSL models.
    \item 
    Combined with i-CutMix, the performance of both PatchSearch and PoisonCAM can be improved, which shows the complementarity of two defense manners.
    \item 
    Our PoisonCAM with i-CutMix consistently achieves the highest ACC with sufficiently low ASR on two datasets, which further verifies the effectiveness of our method in practice.
\end{enumerate}

\begin{figure*}[ht]  
    \begin{minipage}{0.24\linewidth}
        \centerline{\includegraphics[width=\linewidth]{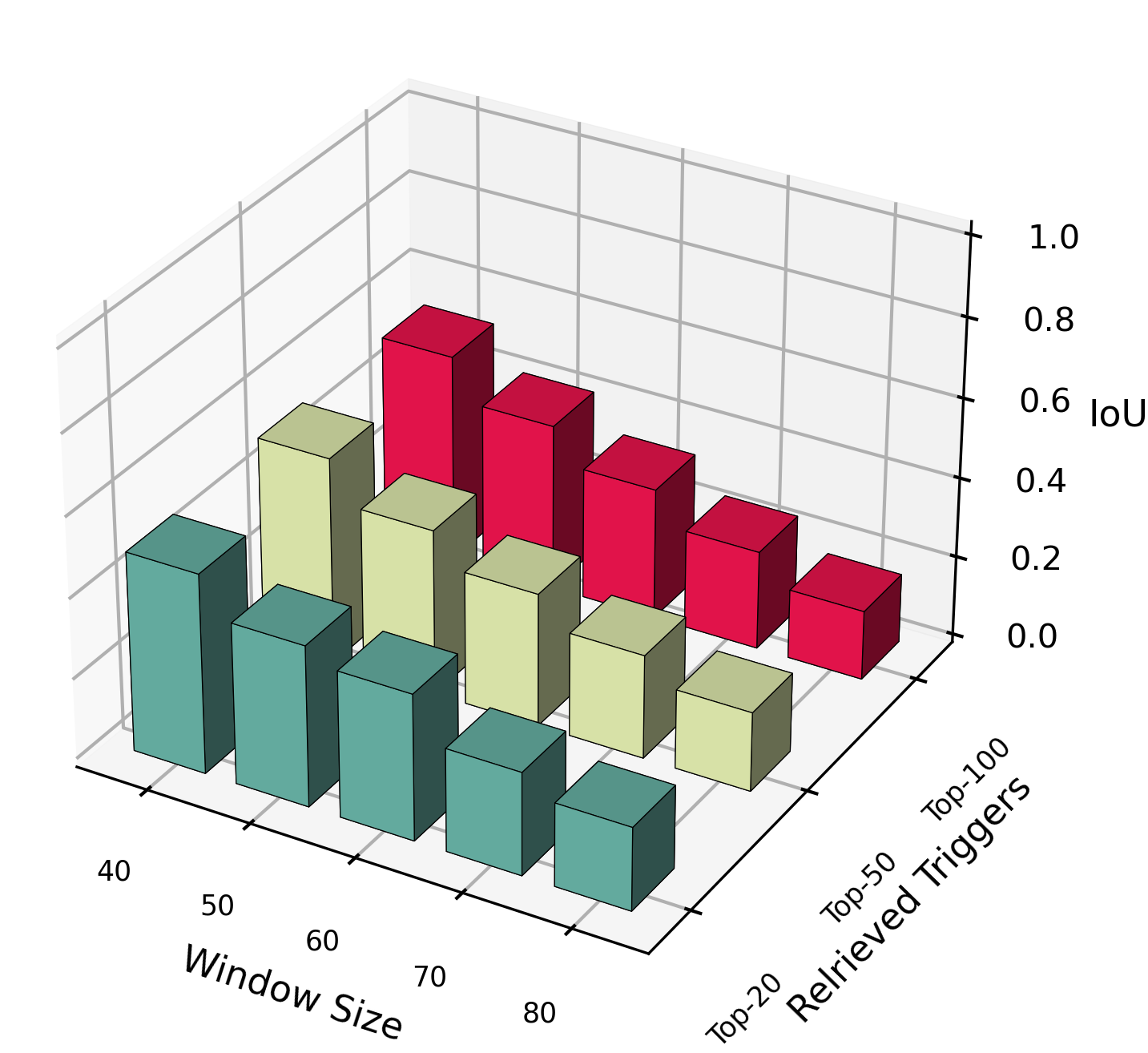}}
        \centerline{\footnotesize{(a) IoU}}
        \label{fig:para_w_a}
    \end{minipage}
    \hfill
    \begin{minipage}{0.24\linewidth}
        \centerline{\includegraphics[width=\linewidth]{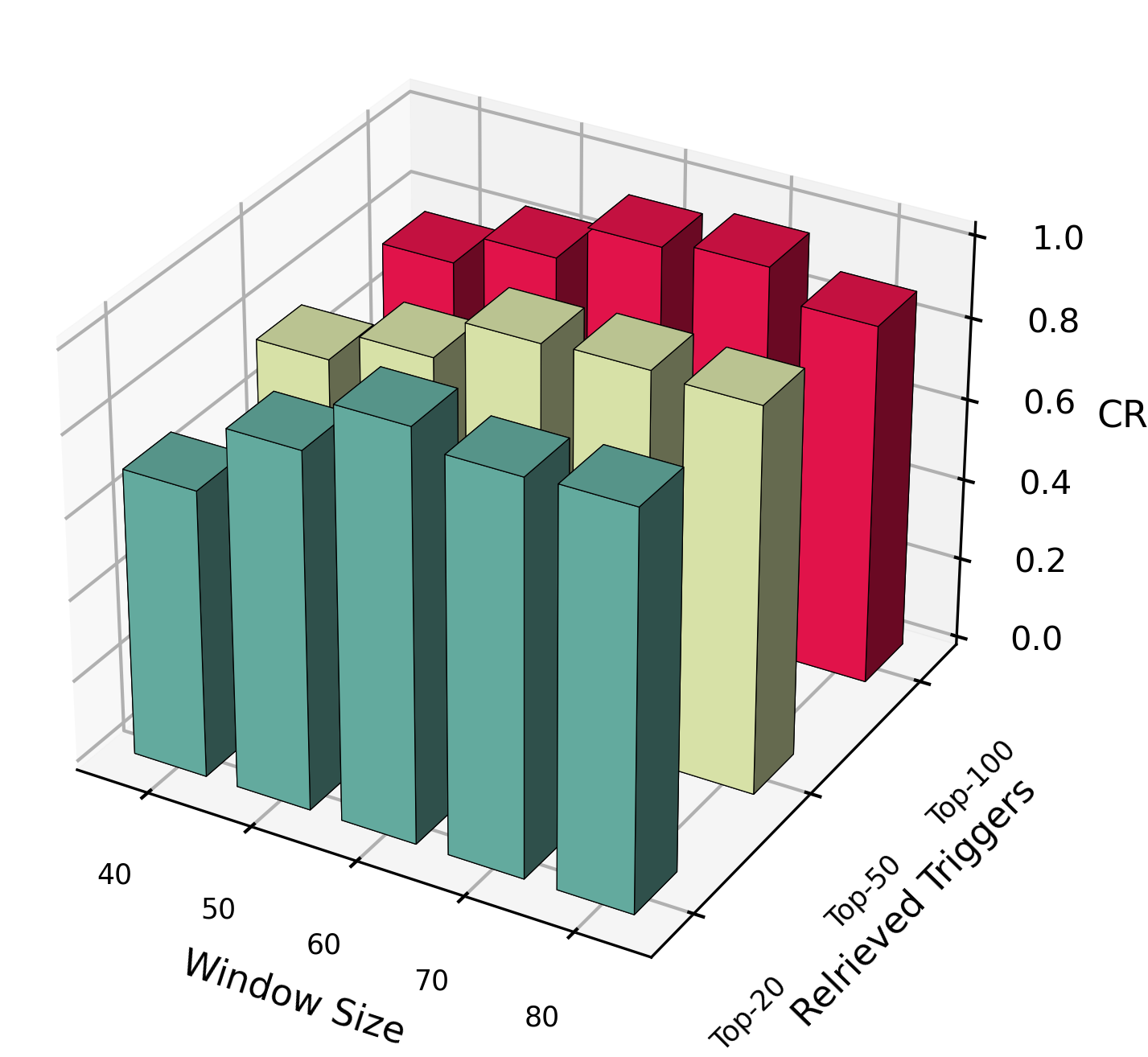}}
        \centerline{\footnotesize{(b) CR}}
        \label{fig:para_w_b}
    \end{minipage}
    \hfill
    \begin{minipage}{0.24\linewidth}
        \centerline{\includegraphics[width=\linewidth]{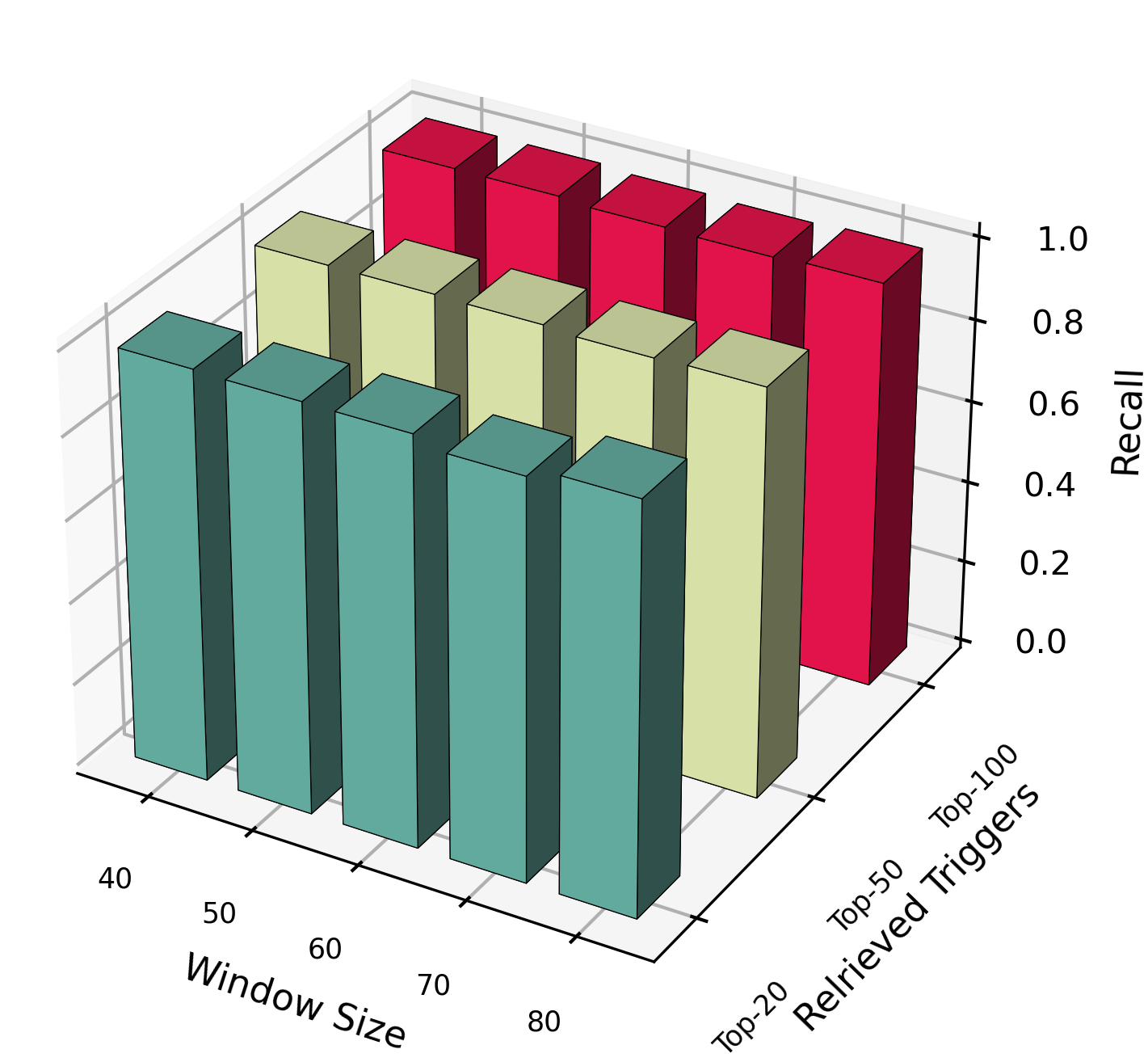}}
        \centerline{\footnotesize{(c) Recall}}
        \label{fig:para_w_c}
    \end{minipage}
    \hfill
    \begin{minipage}{0.24\linewidth}
        \centerline{\includegraphics[width=\linewidth]{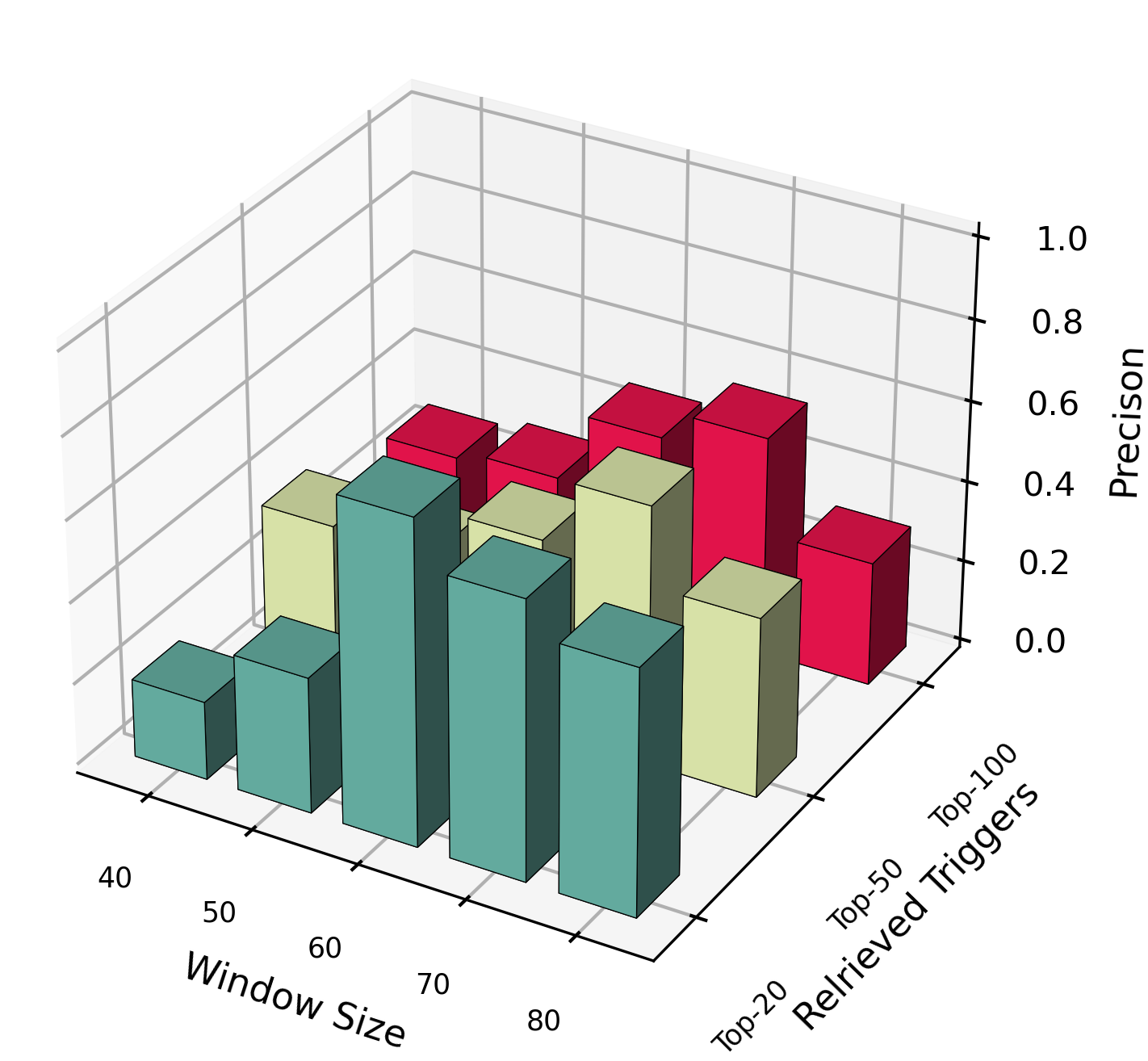}}
        \centerline{\footnotesize{(d) Precision}}
        \label{fig:para_w_d}
    \end{minipage}  
    \vfill
    \caption{Hyper-parameter sensitivity study on the window size $w$ of masks on ImageNet-100 (poison rate 0.5\%, target category ``rottweiler"). (a-b) IoU and CR of the detected candidate triggers. (c-d) Recall and precision of the trained poison classifier.}
    \label{fig:para_w}
\end{figure*} 

\begin{figure*}[ht]  
    \begin{minipage}{0.22\linewidth}
        \centerline{\includegraphics[width=\linewidth]{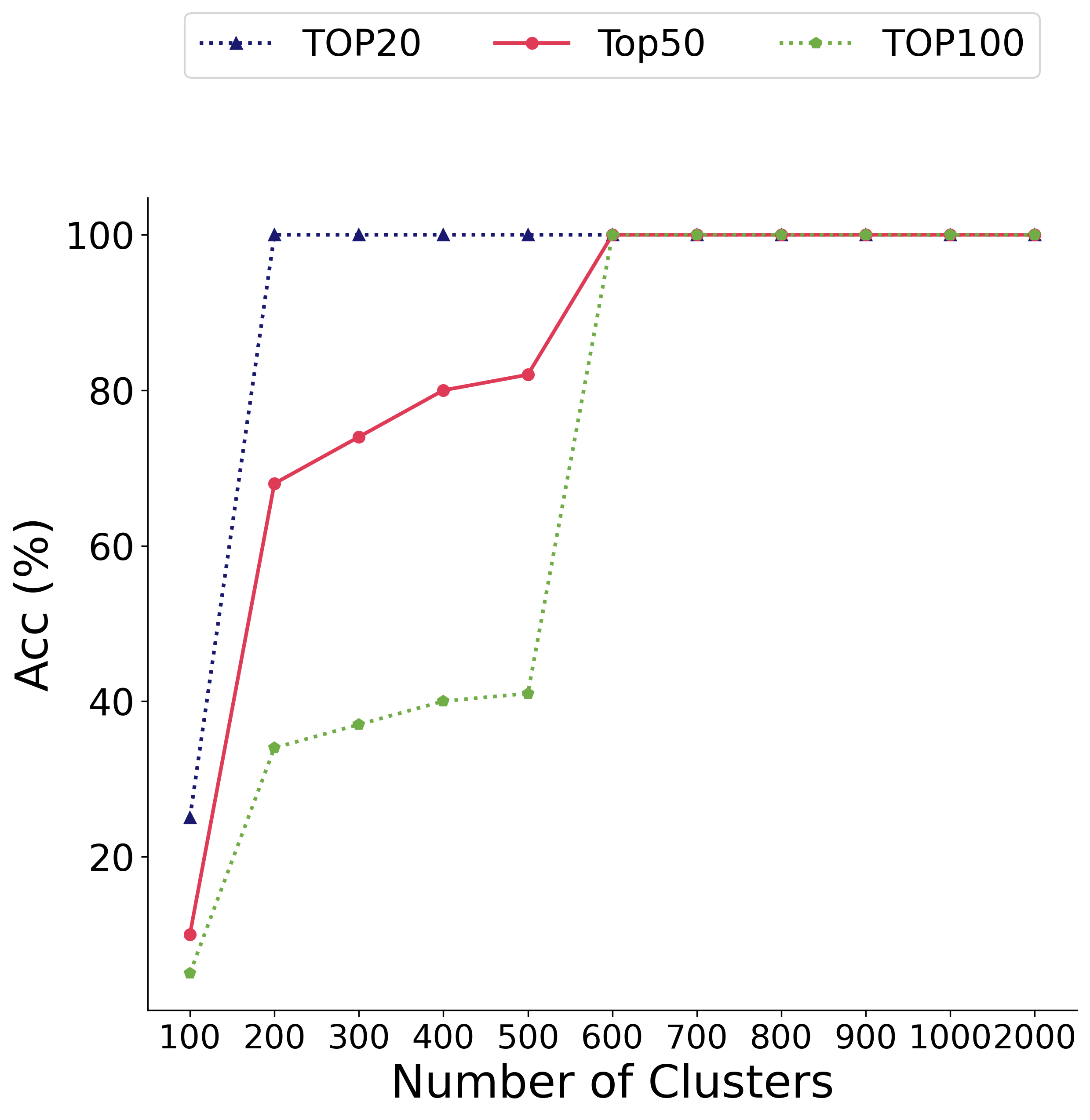}}
        \centerline{\footnotesize{(a) Acc}}
        \label{fig:para_l_a}
    \end{minipage}
    \hfill
    \begin{minipage}{0.22\linewidth}
        \centerline{\includegraphics[width=\linewidth]{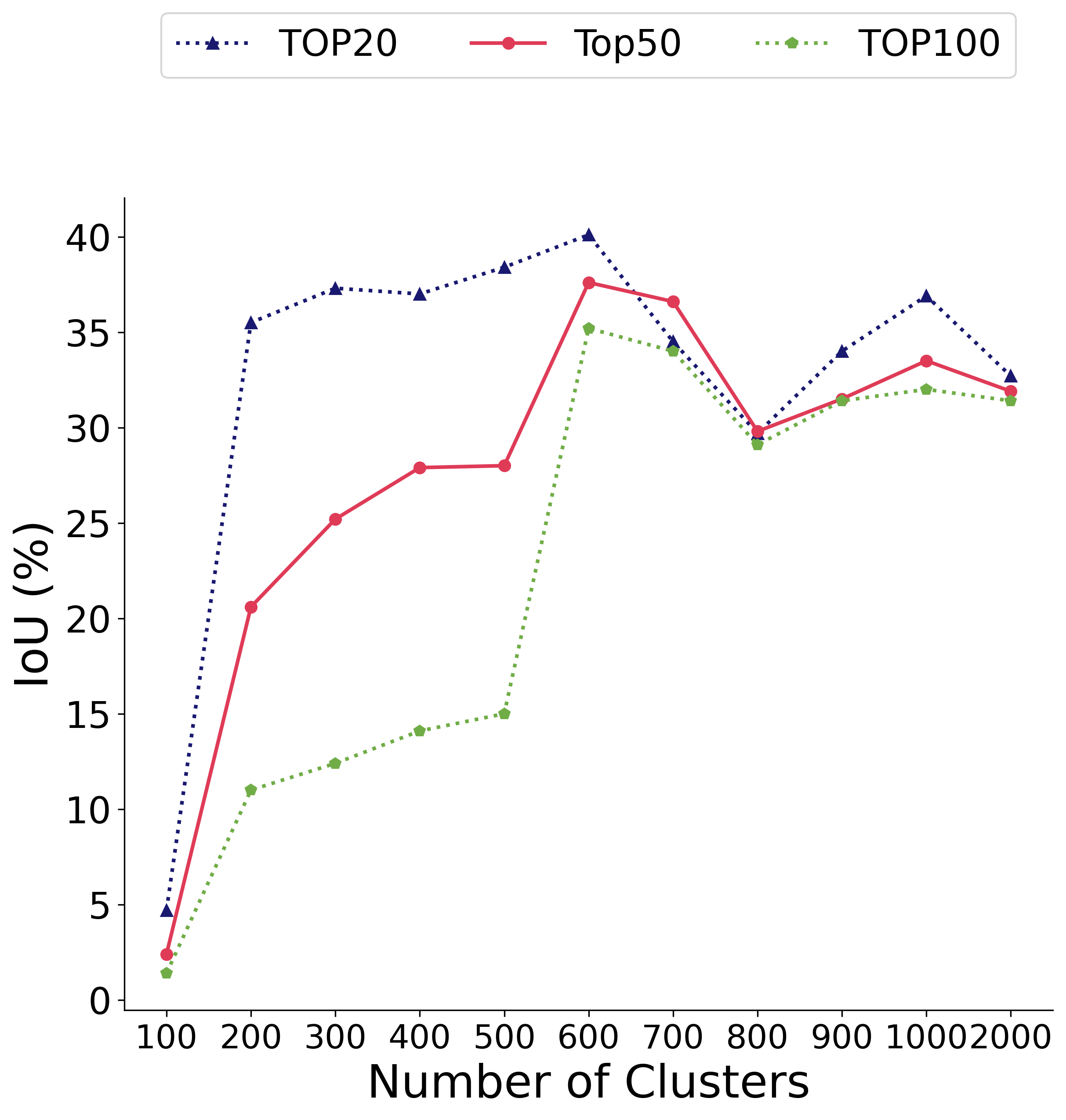}}
        \centerline{\footnotesize{(b) IoU}}
        \label{fig:para_l_b}
    \end{minipage}
    \hfill
    \begin{minipage}{0.22\linewidth}
        \centerline{\includegraphics[width=\linewidth]{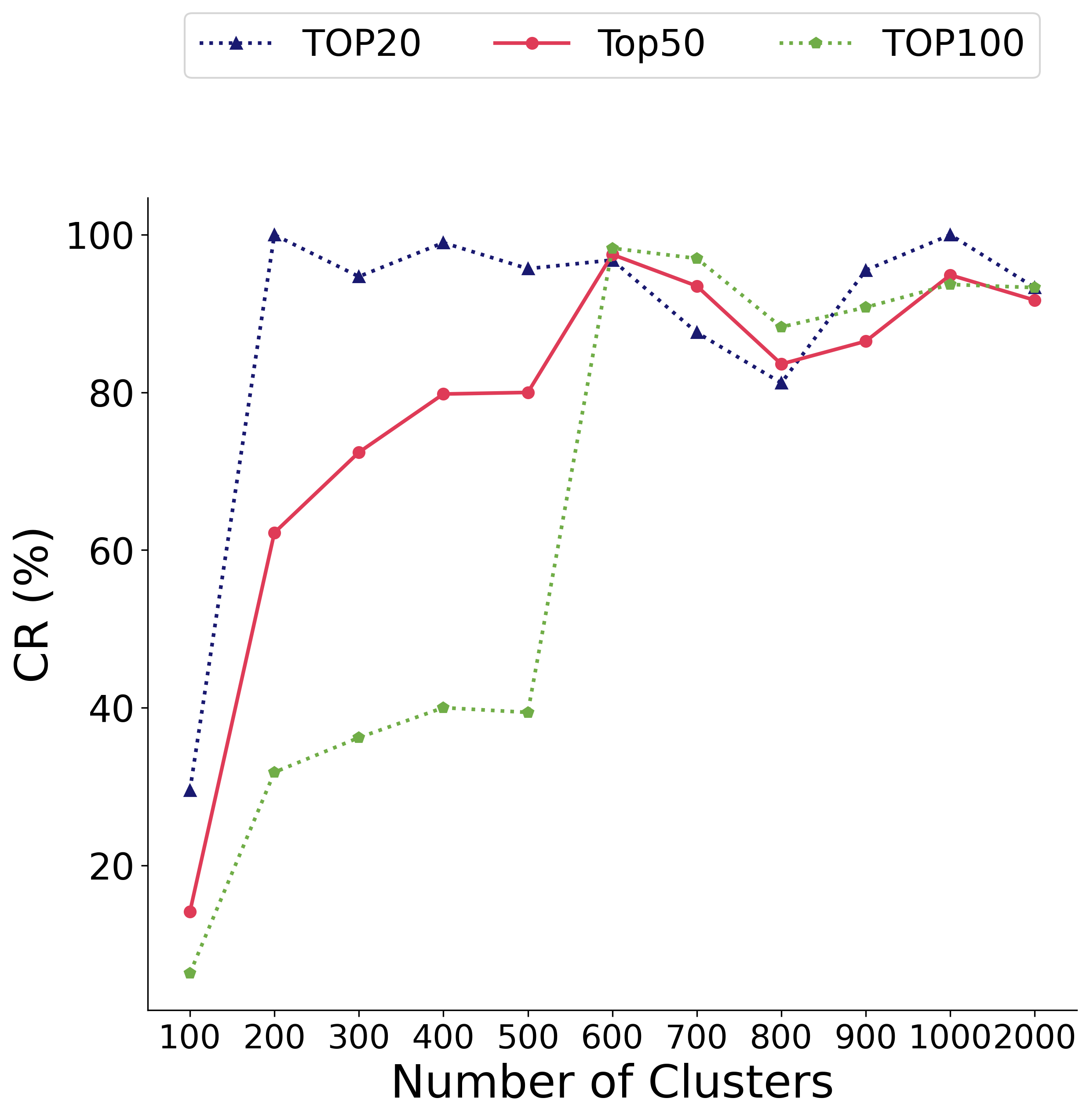}}
        \centerline{\footnotesize{(c) CR}}
        \label{fig:para_l_c}
    \end{minipage}
    \hfill
    \begin{minipage}{0.30\linewidth}
        \centerline{\includegraphics[width=\linewidth]{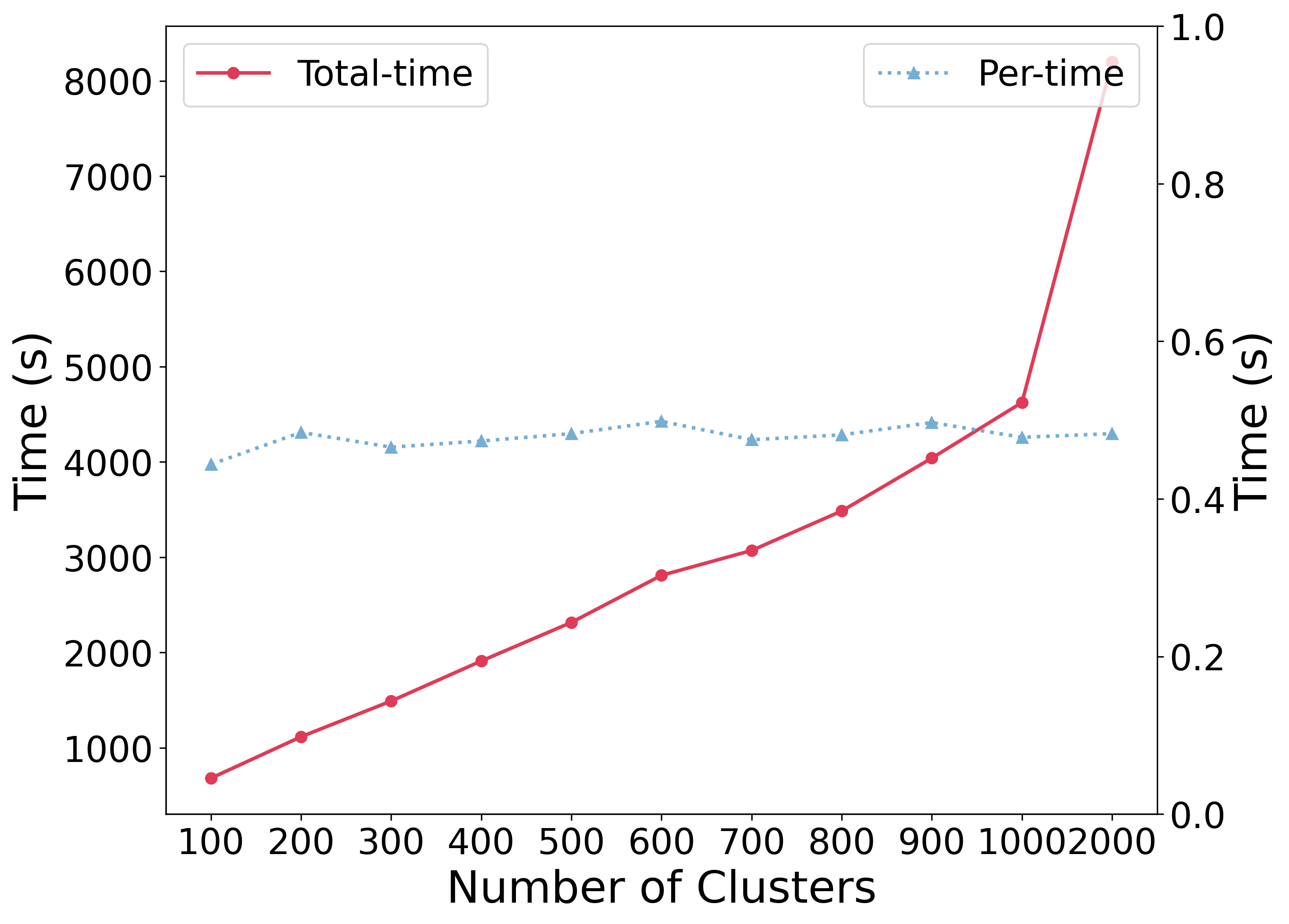}}
        \centerline{\footnotesize{(d) Time}}
        \label{fig:para_l_d}
    \end{minipage}  
    \vfill
    \caption{Hyper-parameter sensitivity study on the cluster count of $l$ on ImageNet-100 (poison rate 0.5\%, target category ``rottweiler"). (a-c) Acc, IoU, and CR of the detected candidate triggers. (d) Total processing time and average processing time per image in candidate trigger detection phase.}
    \label{fig:para_l}
\end{figure*}

\subsection{Hyper-parameter Sensitivity Study} \label{sec:parameter}
In this section, we analyze the hyper-parameter sensitivity on ImageNet-100 (poison rate 0.5\%) with the target category ``rottweiler".

\textbf{Number $B$ of masks.} As demonstrated in Figure \ref{fig:para}, we have the following observations:
(1) While increasing the number $B$ of masks, IoU, CR of the detected candidate triggers and F1-score of the trained poison classifier first increase and then become relatively stable, which shows that our method can effectively locate the triggers with appropriate large $B$. The searching time increases with more masks to calculate.
(2) The average processing time consistently increases while the number $B$ of masks increases. Considering the worst situation where all masks are processed serially, the slope of the time curve will tend towards a constant.
To conclude, an appropriate value of $B$ should be chosen to balance the performance and time costs.
We empirically set $B=256$ in our method.

\textbf{Window size $w$ of masks.} As shown in Figure \ref{fig:para_w}, we vary window size $w$ of masks from 40 to 80, where Acc measures if any part of the trigger is contained in the retrieved window. We have the following observations:
(1) While increasing $w$, the IoU of the detected candidate triggers sharply drops for the reason of more trigger-related regions in candidate triggers.  
(2) CR and Precision first increase and then relatively drop while $w$ increases, which demonstrates that our PoisonCAM can accurately search the triggers with an appropriate $w$.
(3) The recall performs relatively stable, which demonstrates the effectiveness of our method.
We empirically set $w=60$ in our method.

\textbf{Number $l$ of clusters.} As shown in Figure \ref{fig:para_l}, we vary the cluster number $l$ in $k$-means algorithm from 100 to 2000 and have the following observations:
(1) While increasing $l$, the Acc of the detected candidate triggers first increases and then becomes relatively stable, and the IoU and CR increase first and then fluctuate. 
These results demonstrate that a large cluster count of $l$ can detect more candidate images with poisonous triggers.
(2) In the candidate trigger detection phase, the total processing time consistently increases and the average processing time performs relatively stable while the cluster count of $l$ increases.
The system will spend more calculating time with a larger cluster count of $l$.
To conclude, an appropriate value of $l$ should be chosen to balance the performance and time costs. We empirically set $l=1000$.

\subsection{Analysis of Backdoored SSL Process}

In our assessment of the backdoored SSL process on ImageNet-100 with a poison rate of $0.5\%$, we conduct a thorough investigation into the correlation between the overall model performance, as measured by Clean Data Acc, and the attack effectiveness, as measured by the Patched Data FP. As shown in Table \ref{tab:poison_epoch}, we evaluate the attack effectiveness using poisoned data to train the self-supervised model with different training epochs. We test the accuracy and false positive rates of the model on clean and poisoned data after fine-tuning on downstream classification tasks using clean datasets. (poison rate 0.5\%, target category ``laptop" and ``goose"). The results show a strong correlation between the effectiveness of the attack and the model performance before 100 epochs. 
Our findings indicate that in the initial epochs ($epoch\le 40$), attack effectiveness has a strong correlation with overall model performance. Although this correlation is not as stringent in later epochs, the attack effectiveness still maintains its overall magnitude. These observations suggest that, when training a model for defense, achieving convergence may not be necessary. Instead, it is sufficient to train the model until the attack effectiveness is comparable to that of the fully trained model in overall magnitude.

\section{Conclusion}
	In this paper, we propose a novel PoisonCAM method to defend against self-supervised learning (SSL) backdoor attacks. 
        PoisonCAM exhibits a robust capacity to accurately detect and remove poisonous data from a poisoned and unlabeled dataset to facilitate detoxified SSL training.
        We propose a Cluster Activation Masking method to accurately retrieve trigger patches injected into the poisoned dataset. Based on the retrieved trigger patches, an effective poison classifier is trained to distinguish between poisonous and clean data in the training set.
	Extensive experiments on ImageNet-100 and STL-10 demonstrate that PoisonCAM outperforms the state-of-the-art method for defending against SSL backdoor attacks.
	We hope this paper can contribute to the safety of artificial intelligence systems.

\ifCLASSOPTIONcaptionsoff
  \newpage
\fi

\bibliographystyle{IEEEtran}
\bibliography{arxiv}

\end{document}